\documentclass[journal,twoside,web]{ieeecolor}
\usepackage{tmi}
\usepackage{cite}
\usepackage{amsmath,amssymb,amsfonts}
\usepackage{algorithmic}
\usepackage{graphicx}
\usepackage{textcomp}
\usepackage{booktabs}
\usepackage{multirow}
\newcommand{\yr}[1]{\textcolor{black}{#1}}
\def\BibTeX{{\rm B\kern-.05em{\sc i\kern-.025em b}\kern-.08em
    T\kern-.1667em\lower.7ex\hbox{E}\kern-.125emX}}
\begin{document}
\title{MeCaMIL: Causality-Aware Multiple Instance Learning for Fair and Interpretable Whole Slide Image Diagnosis}
\author{Yiran Song, \IEEEmembership{Member, IEEE},
        Yikai Zhang,
        Shuang Zhou,
        Guojun Xiong, \IEEEmembership{Member, IEEE},
        Xiaofeng Yang,
        Nian Wang,
        Fenglong Ma, \IEEEmembership{Member, IEEE},
        Rui Zhang, \IEEEmembership{Member, IEEE},
        and Mingquan Lin*, \IEEEmembership{Member, IEEE}%
\thanks{This research is supported in part by the National Institutes of Health
under Award Number R01CA287413-01. }
\thanks{Y. Song, S. Zhou, and R. Zhang are with the Division of Computational Health Sciences, University of Minnesota, Minneapolis, MN 55455 USA (e-mail: song0760@umn.edu; zhou2219@umn.edu; zhan1386@umn.edu).}
\thanks{Y. Zhang is with the Department of Computer Science and Engineering, University of Minnesota, Minneapolis, MN 55455 USA (e-mail: zhan9191@umn.edu).}
\thanks{G. Xiong is with the Department of Computer Science, Harvard University, Cambridge, MA 02138 USA (e-mail: gjxiong@seas.harvard.edu).}
\thanks{X. Yang is with the Department of Radiation Oncology and Winship Cancer Institute, Emory University, Atlanta, GA 30322 USA (e-mail: xyang43@emory.edu).}
\thanks{N. Wang is with the Advanced Imaging Research Center, University of Texas Southwestern Medical Center, Dallas, TX 75235 USA (e-mail: nian.wang@UTSouthwestern.edu).}
\thanks{F. Ma is with the College of Information Sciences and Technology, The Pennsylvania State University, University Park, PA 16802 USA (e-mail: fenglong@psu.edu).}
\thanks{M. Lin is with the Division of Computational Health Sciences, University of Minnesota, Minneapolis, MN 55455 USA (e-mail: lin01231@umn.edu). *Corresponding author}
}

\maketitle

\begin{abstract}
Multiple instance learning (MIL) has emerged as the dominant paradigm for whole slide image (WSI) analysis in computational pathology, achieving strong diagnostic performance through patch-level feature aggregation. However, existing MIL methods face critical limitations: (1) they rely on attention mechanisms that lack causal interpretability, and (2) they fail to integrate patient demographics (age, gender, race), leading to fairness concerns across diverse populations. \yr{These shortcomings hinder clinical translation, where algorithmic bias can exacerbate health disparities.} We introduce \textbf{\yr{MeCaMIL}}, a causality-aware MIL framework that explicitly models demographic confounders through structured causal graphs. Unlike prior approaches treating demographics as auxiliary features, MeCaMIL employs principled causal inference---leveraging \yr{do-calculus and collider structures}---to disentangle disease-relevant signals from spurious demographic correlations. \yr{Extensive evaluation on three benchmarks demonstrates state-of-the-art performance across CAMELYON16 (ACC/AUC/F1: 0.939/0.983/0.946), TCGA-Lung (0.935/0.979/0.931), and TCGA-Multi (0.977/0.993/0.970, five cancer types). Critically, MeCaMIL achieves superior fairness---demographic disparity variance drops by over 65\% relative reduction on average across attributes, with notable improvements for underserved populations. The framework generalizes to survival prediction (mean C-index: 0.653, +0.017 over best baseline across five cancer types). Ablation studies confirm causal graph structure is essential---alternative designs yield 0.048 lower accuracy and 4.2$\times$ worse fairness. These results establish MeCaMIL as a principled framework for fair, interpretable, and clinically actionable AI in digital pathology.} Code will be released upon acceptance.
\end{abstract}


\section{Introduction}

\IEEEPARstart{I}{n} 
digital pathology, histological sections are digitized into whole slide images (WSIs), which capture detailed information on tissue morphology, cellular structures, and the surrounding microenvironment. Microscopic examination of these slides remains the gold standard for cancer diagnosis~\cite{gurcan2009histopathological,litjens2016deep}, playing a central role in studying disease onset and progression~\cite{chen2021multimodal,yao2020whole,zhu2017wsisa} as well as in the development of targeted therapies~\cite{cornish2012whole,madabhushi2009digital,bollhagen2024highly}. However, interpreting WSIs—often exceeding gigapixel resolution (e.g., $40{,}000 \times 40{,}000$ pixels)—is labor-intensive, creating a pressing demand for automated diagnostic tools. Although artificial intelligence (AI) methods offer the promise of faster analysis, the massive resolution and lack of fine-grained annotations hinder conventional supervised training approaches. To address this, multiple instance learning (MIL) has emerged as a practical solution when only slide-level labels are available. MIL treats each WSI as a bag of instances (image patches) and learns to infer slide-level predictions without requiring patch-level supervision. Nevertheless, existing MIL methods~\cite{dsmil,abmil,dgmil} largely capture correlative patterns between patches and outcomes, rather than causal relationships. This often results in unstable predictions and compromised diagnostic reliability, particularly in heterogeneous patient populations or rare disease settings.

\yr{Recent advances integrate causal inference into medical AI to overcome these limitations~\cite{yang2021causal,li2024causality,wang2021causal,ding2022word}. By modeling cause–effect structures, causal methods improve interpretability and robustness. In whole-slide image (WSI) analysis, IBMIL~\cite{yao2023interventional}, MFC~\cite{chen2025multi}, and CaMIL~\cite{zhang2024camil} apply causal multiple instance learning (MIL) to mitigate spurious correlations and enhance pathological interpretability. However, these works primarily address image-based confounders and overlook demographic bias—factors such as age, race, and sex can systematically skew performance across patient subgroups~\cite{seyyed2024demographic, lin2024improving, lin2023improving, lin2023evaluate}. This underscores the need for fairness-aware causal MIL frameworks that explicitly incorporate demographic information and rigorously evaluate performance across diverse patient populations.}

\begin{figure*}[!ht]
\centering
\includegraphics[width=0.9\textwidth]{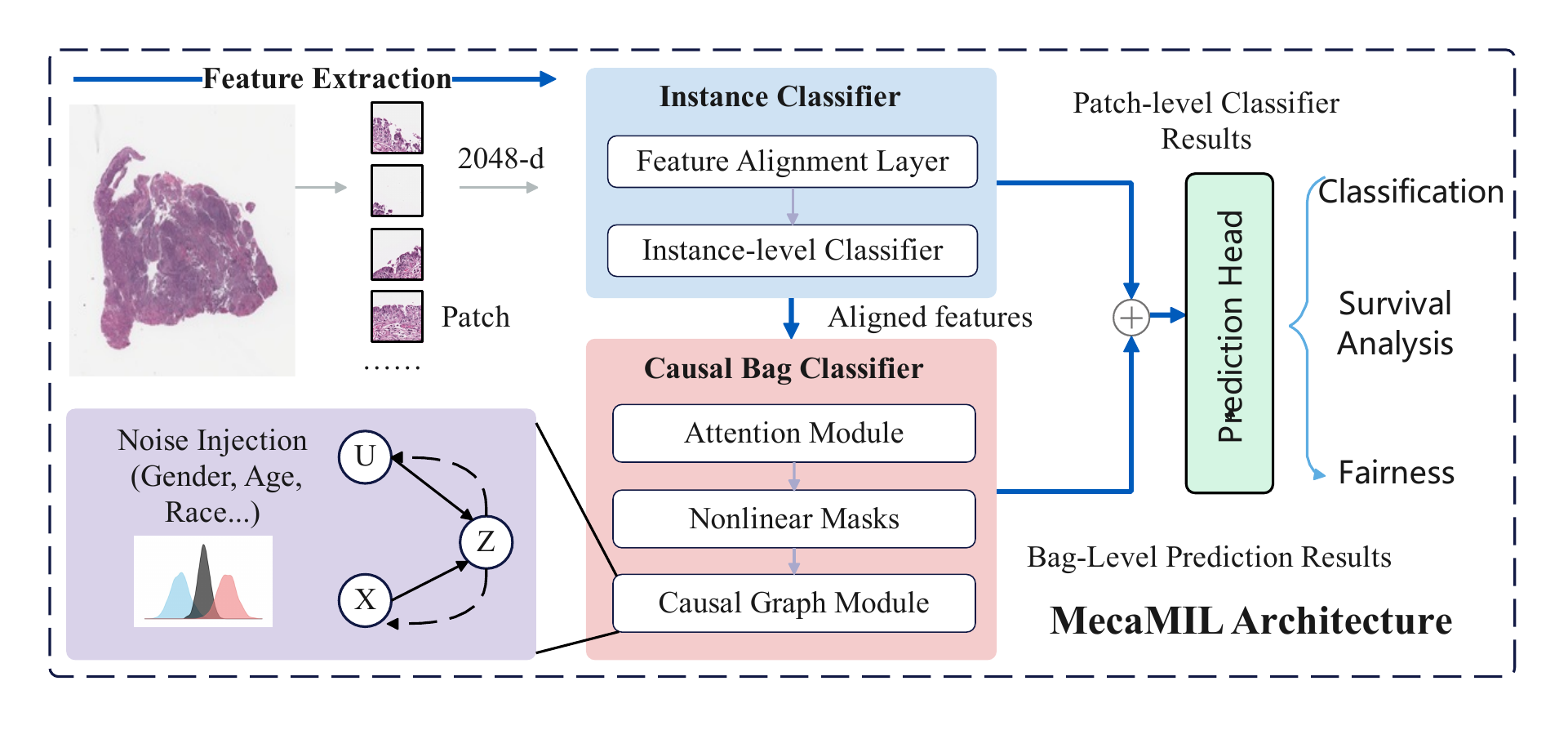}
\caption{Overview of the proposed \textbf{MeCaMIL} for whole slide image (WSI) diagnosis. 
Each WSI is divided into image patches, which are independently encoded into 2048-dimensional instance features using a pretrained encoder. 
These features are first passed to the \textbf{Instance Classifier}, which comprises a Feature Alignment Layer and a shallow Instance-Level Classifier, generating per-patch predictions and aligned feature representations.
In parallel, features are aggregated in the \textbf{Causal Bag Classifier} via a multi-head Attention Module to produce a bag-level embedding. 
This embedding is further processed through Nonlinear Masking Blocks and a Causal Graph Module to capture structured dependencies and inject exogenous demographic priors (e.g., gender, race, age) for debiasing. \yr{In the Causal Graph Module, solid arrows denote causal edges that persist during both training and inference, while dashed arrows represent auxiliary edges used exclusively for gradient backpropagation during training.}
The resulting representations from both instance and bag branches are integrated in the \textbf{Prediction Head}, supporting diverse downstream tasks including classification, survival prediction, and fairness-aware modeling.
Arrows denote the data flow and transformation at each module, with exogenous priors influencing the causal path through latent variable modulation.}

\label{fig_main}
\end{figure*}

To address these gaps, as illustrated in Figure~\ref{fig_main}, we propose \yr{MeCaMIL—a causality-aware multiple instance learning framework that extends beyond traditional bag-level modeling by integrating structured causal reasoning with demographic awareness}. Our architecture offers \yr{two key innovations over existing approaches. First, unlike IBMIL and MFC which primarily target image-level confounders, MeCaMIL explicitly models the causal pathways involving both pathological features and demographic attributes through a graph-based structural equation model. Second, we introduce an exogenous variable injection mechanism that incorporates patient demographic information (age, race, gender) as principled causal interventions rather than simple feature concatenation, enabling systematic bias mitigation}. This design provides a theoretically grounded framework for interpretable and fair diagnosis, contrasting with conventional CNN- or attention-based MIL approaches that rely solely on statistical associations.

Our contributions are fourfold:
\begin{itemize}
\item We propose MeCaMIL, a causality-aware MIL framework that \yr{jointly models pathological image features and demographic factors through graph-based causal structures}, enabling interpretable diagnostic decisions \yr{that explicitly account for both morphological and demographic confounders}.
\item We introduce an exogenous variable injection mechanism to incorporate demographic information (e.g., age, race, gender) as structured \yr{causal interventions, providing a principled approach to} reduce spurious correlations and \yr{promote} fairness \yr{across patient subgroups}.
\item We conduct extensive experiments on CAMELYON16 and TCGA datasets, demonstrating \yr{competitive} classification performance\yr{,} \yr{with} significantly reduced \yr{demographic} disparity, and \yr{further validate the framework's generalizability through} survival prediction tasks.
\end{itemize}

\section{RELATED WORKS}

\subsection{Multiple-Instance Learning for Whole Slide Images}
\yr{MIL frameworks for WSI analysis can be broadly categorized by their aggregation strategies.} ABMIL~\cite{ilse2018attention} marked a pivotal shift by introducing attention\yr{-based instance weighting} \yr{mechanisms to weight instance contributions}, inspiring subsequent refinements such as cosine-similarity weighting~\cite{li2021dual} \yr{and dual-stream architectures~\cite{dsmil}}. More recent methods \yr{leverage} transformer architectures~\cite{transmil,vaswani2017attention} to capture \yr{long-range} intra-slide \yr{interactions}, while others~\cite{lu2021data} incorporate clustering \yr{constraints} to reduce instance ambiguity. Despite these advances, most existing MIL methods focus on learning correlative patterns and do not explicitly address confounding factors or spurious correlations, which can undermine model reliability in heterogeneous clinical populations.

\subsection{Causal Inference in Whole Slide Image Analysis}
Causal inference provides principled tools to identify and mitigate confounding through interventional strategies grounded in structural causal models~\cite{pearl2016causal,pearl2014interpretation}. \yr{Recent work has begun integrating causal reasoning into WSI analysis to improve robustness and interpretability.} IBMIL~\cite{yao2023interventional} \yr{introduces} backdoor adjustment at the bag level to eliminate confounders, achieving improved generalization by breaking spurious feature-label associations. \yr{MFC~\cite{chen2025multi} leverages multi-scale frequency domain analysis within a causal framework to disentangle pathology-relevant signals from noise. CaMIL~\cite{zhang2024camil} extends causal modeling to the instance level, capturing fine-grained dependencies among patches for more accurate slide classification.}
While these methods demonstrate the value of causal reasoning in computational pathology, they primarily target image-based confounders and do not systematically incorporate patient-level attributes such as demographics. Moreover, they lack mechanisms for evaluating or enforcing fairness across sensitive subgroups—an increasingly critical concern in clinical AI deployment.

\subsection{Fairness in Computational Pathology}
Algorithmic fairness has emerged as a pressing issue in medical AI, as deep learning models have been shown to exhibit systematic performance disparities across subgroups defined by sex, race, age, or socioeconomic status~\cite{stanley2022fairness,seyyed2021underdiagnosis}. In computational pathology, recent studies~\cite{seyyed2024demographic} reveal that AI-based diagnostic systems can produce biased predictions due to imbalanced training data or spurious correlations between demographic attributes and histological patterns. These biases undermine key bioethical principles~\cite{beauchamp2003methods} and pose significant barriers to equitable clinical adoption~\cite{liu2023translational}.
Traditional fairness metrics—such as demographic parity and equalized odds—focus on statistical independence between predictions and sensitive attributes but often fail to address underlying causal mechanisms~\cite{srivastava2019mathematical}. Without causal insight, \yr{mitigation strategies may be unreliable or even introduce new disparities}. Moreover, fairness is inherently context-dependent and evolves with societal values~\cite{jones2010building}, necessitating flexible frameworks that can adapt to different clinical settings. Addressing fairness in computational pathology thus requires moving beyond correlation-based approaches toward causal frameworks that can reveal and intervene on the root causes of bias. By explicitly modeling the causal pathways linking demographic factors to diagnostic outcomes, such frameworks can support more principled fairness evaluation and mitigation strategies.

\section{BACKGROUND}
In this section, we provide the theoretical foundations for causal inference and multiple instance learning that underpin our framework.

\subsection{Causal Inference Framework}
\label{sec:causal_framework}
The framework for causal inference consists of three main components: causal graphs to represent variable relationships, causal models to identify causal effects, and probability expressions for quantification.

\textbf{Causal Graph}: A causal graph is a \yr{directed acyclic graph (DAG) that encodes causal relationships among variables}. For a set of variables $V$, if variable $V_i$ \yr{directly affects} $V_j$, then $V_i$ is called the parent of $V_j$. \yr{Five fundamental structures serve as building blocks:}

\begin{enumerate}
\item \textbf{Chain}: $X \rightarrow W \rightarrow Y$ — $X$ and $Y$ are statistically dependent\yr{; conditioning on $W$ blocks the path}.
\item \textbf{Fork}: $X \leftarrow W \rightarrow Y$ — $X$ and $Y$ are dependent\yr{; conditioning on $W$ blocks the path}.
\item \textbf{Collider}: $X \rightarrow W \leftarrow Y$ — $X$ and $Y$ are independent\yr{; conditioning on $W$ \textit{opens} the path}.
\item \textbf{Two Unconnected Nodes}: $X \quad Y$ — statistically independent.
\item \textbf{Two Connected Nodes}: $X \rightarrow Y$ — statistically dependent.
\end{enumerate}

\textbf{Causal Model}: A causal model is formally defined as a triple $M = \langle U, V, F \rangle$ where:
\begin{itemize}
\item $U$ is the set of \textit{exogenous} variables (determined by factors outside the model);
\item $V$ is the set of \textit{endogenous} variables (determined within the model by $U \cup V$ \yr{through functions in $F$});
\item $F$ is a set of \yr{structural equations}, where each $V_i \in V$ is determined by its parents $pa_i \subseteq V \setminus \{V_i\}$ and corresponding exogenous \yr{noise} $U_i$:
\end{itemize}

\begin{equation}
V_i = F_i(pa_i, U_i)
\end{equation}

For example, consider $M = \langle U, V, F \rangle$ with $V = \{V_1, V_2\}$, $U = \{U_1, U_2, I, J\}$, and \yr{structural equations}:
\begin{align}
V_1 &= F_1(I, U_1) = \theta_1 I + U_1 \\
V_2 &= F_2(V_1, J, U_2) = \phi V_1 + \theta_2 J + U_2
\end{align}

\yr{The \textbf{do-operator} $do(X=x)$ represents external interventions that set variable $X$ to value $x$, enabling computation of causal effects via interventional distributions $P(Y|do(X=x))$, which differ from observational distributions $P(Y|X=x)$ when confounders are present.}

\subsection{Multiple Instance Learning Framework}

\textbf{Problem Formulation}: Consider a binary classification \yr{task} with a dataset $D = \{(X_1, Y_1), \ldots, (X_n, Y_n)\}$. Each input $X_i$ is a bag containing instances $\{x_1, \ldots, x_K\}$, with bag label $Y_i \in \{0, 1\}$. Individual instance labels $\{y_1, \ldots, y_K\}$ are unobserved during training. The bag label is determined by:

\begin{equation}
Y = \max_{i} y_i
\end{equation}

\yr{where a bag is positive if at least one instance is positive.}

\yr{\textbf{MIL Paradigms}: According to the classification granularity and feature space where learning occurs, MIL methods can be categorized into three main paradigms~\cite{amores2013multiple}:}

\begin{enumerate}
\item \textbf{Instance-space methods}: Learn instance-level classifiers to predict individual instance labels, then aggregate instance predictions (e.g., via max-pooling) to infer the bag label. While interpretable at the instance level, these methods may struggle when the standard MIL assumption does not hold strictly.

\item \textbf{Bag-space methods}: Treat each bag as a holistic entity and define distance or similarity measures between bags for classification. These methods directly compare bags without explicitly modeling individual instances, offering robustness but limited interpretability.

\item \textbf{Embedding-space methods}: Map each bag to a fixed-dimensional feature vector through instance feature aggregation (e.g., attention-based pooling), then apply standard classifiers. Modern attention-based MIL methods~\cite{ilse2018attention} fall into this category, achieving superior performance by learning adaptive instance weighting.
\end{enumerate}

\yr{Our proposed framework follows the embedding-space paradigm, using attention mechanisms to aggregate instance features into bag-level representations while incorporating causal modeling to address demographic confounders.}

\section{METHODOLOGY}

This section presents MeCaMIL, a causality-aware multiple instance learning framework that addresses fairness and interpretability in WSI diagnosis through explicit causal modeling of demographic confounders.

\subsection{Problem Formulation and Causal Framework}

\subsubsection{Causal Graph Design}

We model WSI diagnosis using a DAG with a \textit{collider} structure where image features $X$ and demographic variables $U$ independently influence latent disease representation $Z$, which determines diagnosis $Y$:

\begin{equation}
X \rightarrow Z \leftarrow U, \quad Z \rightarrow Y
\label{eq:causal_structure}
\end{equation}

\yr{\textbf{Connection to Fundamental Structures}: Our causal graph combines two basic structures: (1) a \textbf{collider} at node $Z$ where $X \rightarrow Z \leftarrow U$, and (2) a \textbf{chain} from $X$ through $Z$ to $Y$ (i.e., $X \rightarrow Z \rightarrow Y$), with an analogous chain from $U$ to $Y$. This design ensures that:}
\begin{itemize}
\item $X$ and $U$ are marginally independent (blocked at collider $Z$);
\item Both $X$ and $Y$, and $U$ and $Y$, are dependent through causal chains;
\item Conditioning on $Z$ opens the collider, allowing information flow from both sources.
\end{itemize}

\yr{\textbf{Clinical Rationale}: This structure is motivated by domain knowledge: (1) pathological image features $X$ directly reflect tissue-level disease manifestation; (2) demographic attributes $U$ (age, race, gender) influence disease presentation through distinct biological pathways (e.g., hormonal differences, genetic predispositions); (3) both converge to determine the underlying disease state $Z$, which then causally determines the diagnosis $Y$. This differs from a fork structure ($Z \rightarrow X, Z \rightarrow U$), which incorrectly suggests disease state causes demographics, and avoids direct $U \rightarrow Y$ edges that would conflate demographic bias with true causal effects.}

Our structural causal model formally defines:

\begin{align}
X &= f_X(\epsilon_X), \quad U = f_U(\epsilon_U) \quad \text{(exogenous)} \label{eq:exogenous}\\
Z &= f_Z(X, U, \epsilon_Z) \quad \text{(disease representation)} \label{eq:disease_rep}\\
Y &= f_Y(Z, \epsilon_Y) \quad \text{(diagnosis)} \label{eq:diagnosis}
\end{align}

where $\epsilon_{\cdot}$ denote independent noise terms and $f_{\cdot}$ are neural network functions. \yr{This enables causal effect quantification via interventional distributions $P(Y|do(U=u))$.}

\subsubsection{Identifiability Assumptions}

\yr{Our causal inference framework relies on three standard assumptions: (1) \textbf{Conditional Independence}: potential outcomes under different demographic interventions are independent of actual demographics given disease state, $(Y^{u=0}, Y^{u=1}) \perp U | Z$, which is enforced architecturally by the collider structure; (2) \textbf{Positivity}: all combinations of $(X,U)$ have positive probability in the population; (3) \textbf{Consistency}: observed outcomes equal potential outcomes under observed conditions, i.e., $Y = Y^{u=U}$.}

\subsection{Architecture Design}

\subsubsection{Feature Extraction and Weakly-Supervised Instance Classification}

Given a WSI represented as a bag $\mathcal{B} = \{x_1, ..., x_K\}$ where each patch $x_i \in \mathbb{R}^{H \times W \times 3}$, we first extract patch-level embeddings using a pre-trained encoder:

\begin{equation}
h_i = \Phi(x_i) \in \mathbb{R}^d
\label{eq:feature_extraction}
\end{equation}

where $\Phi(\cdot)$ is a ResNet-18 pre-trained on ImageNet with output dimension $d=512$.

\yr{\textbf{Weakly-Supervised Instance Prediction}: Since patch-level annotations are unavailable, we train an instance classifier using only bag-level labels through a pseudo-labeling strategy. Each patch embedding is passed through a fully connected layer:}

\begin{equation}
c_i = \text{FC}(h_i) \in \mathbb{R}^C
\label{eq:instance_classifier}
\end{equation}

\yr{where $C$ is the number of classes. During training, we assign pseudo-labels to instances based on the bag label: for positive bags ($Y=1$), we assign pseudo-label 1 to the top-$k$ instances with highest prediction scores, and label 0 to the remaining instances; for negative bags ($Y=0$), all instances receive label 0. The instance classification loss is:}

\begin{equation}
\mathcal{L}_{\text{ins}} = \frac{1}{K}\sum_{i=1}^K \text{CE}(c_i, \tilde{y}_i)
\label{eq:instance_loss_detail}
\end{equation}

\yr{where $\tilde{y}_i$ are the pseudo-labels and CE denotes cross-entropy loss. This provides weak supervision that guides the model to identify diagnostically relevant patches.}

\subsubsection{Causality-Aware Attention Mechanism}

Our attention mechanism identifies diagnostically critical instances through a query-based approach:

\begin{align}
j^* &= \arg\max_j c_j, \quad h_{\text{crit}} = h_{j^*} \label{eq:critical_instance}\\
Q_i &= \tanh(W_q h_i), \quad Q_{\text{crit}} = \tanh(W_q h_{\text{crit}}) \label{eq:query_transform}\\
\alpha_i &= \frac{\exp(Q_i^T Q_{\text{crit}} / \sqrt{d_q})}{\sum_{j=1}^K \exp(Q_j^T Q_{\text{crit}} / \sqrt{d_q})} \label{eq:attention_weights}\\
B &= \sum_{i=1}^K \alpha_i h_i \label{eq:bag_representation}
\end{align}

\yr{Here, $h_{\text{crit}}$ is the embedding of the most confident instance, serving as a query to identify similar instances.} The resulting bag representation $B \in \mathbb{R}^d$ serves as the image input $X$ in our causal model.

\subsubsection{Graph Neural Network-Based Structural Equation Model}

To implement the causal relationship $Z = f_Z(X, U, \epsilon_Z)$ from Eq.~\ref{eq:disease_rep}, we employ a graph attention network (GAT) that explicitly models causal information flow according to our DAG structure.

\yr{\textbf{Node Initialization}: Each causal variable is embedded into a hidden space $\mathbb{R}^{d_h}$ ($d_h=256$):}

\begin{align}
h_X &= \text{Dropout}(\text{GELU}(\text{LayerNorm}(\text{Linear}(B)))) \label{eq:embed_x}\\
h_U &= \text{Dropout}(\text{GELU}(\text{LayerNorm}(\text{Linear}(u)))) \label{eq:embed_u}\\
h_Z^{(0)} &= \text{Linear}(\mathbf{0}_{d_h}) \label{eq:init_z}
\end{align}

\yr{where $u \in \mathbb{R}^{8}$ is a one-hot encoding of demographics: gender (2 dimensions), race (5 dimensions), and age group (1 dimension). The node $h_Z$ is initialized with zeros, as it is endogenous and determined by its parents.}

\yr{\textbf{Graph Structure and the Graph($\cdot$) Function}: The $\text{Graph}(\cdot)$ function in our framework implements message passing over the causal DAG through a structured adjacency matrix $\mathbf{A}$:}

\begin{equation}
\mathbf{A} = \begin{bmatrix}
1 & 0 & 1 \\
0 & 1 & 1 \\
1 & 1 & 1
\end{bmatrix} \quad 
\begin{array}{l}
\text{Row 1: X node (self-loop, edge to Z)} \\
\text{Row 2: U node (self-loop, edge to Z)} \\
\text{Row 3: Z node (receives from X, U, self)}
\end{array}
\label{eq:adjacency}
\end{equation}

\yr{The Graph($\cdot$) operation is implemented as a masked multi-head graph attention layer:}

\begin{align}
&\text{Graph}([h_X, h_U, h_Z], \mathbf{A}) = [h_X', h_U', h_Z'] \label{eq:graph_function}\\
&\text{where} \quad h_i' = h_i + \sum_{j: \mathbf{A}_{ij}=1} \alpha_{ij} W_v h_j \label{eq:graph_update}\\
&\alpha_{ij} = \frac{\exp(\text{score}_{ij})}{\sum_{k: \mathbf{A}_{ik}=1} \exp(\text{score}_{ik})}, \quad \text{score}_{ij} = \frac{(W_q h_i)^T (W_k h_j)}{\sqrt{d_h/n_h}} \label{eq:graph_attention}
\end{align}

\yr{Here, $W_q, W_k, W_v$ are learnable projection matrices, $n_h=4$ is the number of attention heads, and the adjacency mask ensures information flows only along valid causal edges. Specifically:}
\begin{itemize}
\item \yr{Node $X$ attends to itself (preserving image information);}
\item \yr{Node $U$ attends to itself (preserving demographic information);}
\item \yr{Node $Z$ attends to both $X$ and $U$ (implementing the collider structure $X \rightarrow Z \leftarrow U$).}
\end{itemize}

\yr{\textbf{Bidirectional Edges for Training}: While our causal graph is directed $X \rightarrow Z \leftarrow U$ for inference, we include reverse edges $Z \rightarrow X$ and $Z \rightarrow U$ in $\mathbf{A}$ during training to enable gradient flow. This does not violate causal semantics because: (1) during forward inference, we only use the final $h_Z$ to predict $Y$, ignoring the updated $h_X'$ and $h_U'$; (2) the reverse edges serve purely as a mechanism for backpropagation, allowing the model to learn representations that satisfy the causal constraints.}

After $L=1$ message passing layer, the updated disease representation $h_Z^{(1)}$ encodes causal effects from both image features and demographics:

\begin{equation}
Z = h_Z^{(1)} = \text{Graph}([h_X, h_U, h_Z^{(0)}], \mathbf{A})[2]
\label{eq:final_z}
\end{equation}

where $[\cdot][2]$ denotes selecting the third element (node $Z$) from the output node list.

\yr{\textbf{Role of Demographic Encoding $Z_u$}: In our implementation, $Z_u := h_U$ represents the embedded demographic information that directly influences the disease representation $Z$ through the graph message passing. The alignment of $Z_u$ with demographic labels serves two purposes: (1) it ensures the model learns meaningful demographic embeddings that encode relevant attributes (age, race, gender); (2) it enables demographic disentanglement—by explicitly reconstructing demographics from $Z$, we encourage the model to preserve demographic information in a structured manner.
Formally, we add a demographic reconstruction loss:}

\begin{equation}
\mathcal{L}_{\text{demo}} = ||D_{\text{dec}}(Z) - u||_2^2
\label{eq:demo_reconstruction}
\end{equation}

\yr{where $D_{\text{dec}}(\cdot)$ is a decoder network (2-layer MLP). This loss ensures that $Z$ retains demographic information, which is critical for interpretability—we can attribute prediction changes to specific demographic factors.}

\yr{\textbf{Handling Missing Demographics}: When demographic data is incomplete, we use a learned imputation mechanism:}


\begin{equation}
u_{\text{final}} = \begin{cases}
\sigma_{\text{unc}} \cdot \text{MLP}_{\text{impute}}(B) & \text{if $u$ is fully} \\
& \text{missing} \\
\mathcal{M} \odot u + (1-\mathcal{M}) \odot \text{MLP}_{\text{impute}}(B) & \text{if $u$ is partially} \\
& \text{observed}
\end{cases}
\label{eq:demo_handling}
\end{equation}

\yr{where $\mathcal{M} \in \{0,1\}^8$ is the missingness indicator, $\text{MLP}_{\text{impute}}$ predicts missing attributes from image features, and $\sigma_{\text{unc}} < 1$ is an uncertainty weight that reduces reliance on imputed values.}





\subsubsection{Causal Attribution Analysis}

\yr{To quantify the influence of demographics on predictions, we employ intervention-based attribution:}

\begin{align}
\text{Attribution}_U &= ||Z(X, U) - Z(X, U=\mathbf{0})||_2 \label{eq:total_attribution}\\
\text{Attribution}_{u_j} &= ||Z(X, U) - Z(X, U_{-j})||_2 \label{eq:factor_attribution}
\end{align}

\yr{where $U_{-j}$ denotes demographics with the $j$-th attribute set to a neutral value (e.g., population mean). $\text{Attribution}_U$ measures total demographic influence, while $\text{Attribution}_{u_j}$ isolates the effect of individual factors (e.g., age, race, gender). These metrics provide interpretable causal explanations for model decisions.}

\subsection{Training Objectives}

The disease representation $Z$ is projected to diagnostic logits via a 1D convolution:

\begin{equation}
\hat{y}_{\text{bag}} = \text{Conv1D}(Z^T) \in \mathbb{R}^C
\label{eq:bag_prediction}
\end{equation}

Our training objective combines multiple loss terms:

\begin{align}
\mathcal{L}_{\text{total}} &= \mathcal{L}_{\text{cls}} + \lambda_{\text{causal}}\mathcal{L}_{\text{causal}} \yr{+ \lambda_{\text{fair}}\mathcal{L}_{\text{fair}}} \label{eq:total_loss}
\end{align}

\textbf{Classification Loss}: Combines bag-level and instance-level supervision:
\begin{equation}
\mathcal{L}_{\text{cls}} = \text{CE}(\hat{y}_{\text{bag}}, y_{\text{bag}}) + \lambda_{\text{ins}}\mathcal{L}_{\text{ins}}
\label{eq:classification_loss}
\end{equation}

where $\mathcal{L}_{\text{ins}}$ is defined in Eq.~\ref{eq:instance_loss_detail}.

\yr{\textbf{Causal Consistency Loss}: Encourages the disease representation to remain close to the original image features (preventing over-reliance on demographics) while preserving demographic information:}
\begin{equation}
\mathcal{L}_{\text{causal}} = ||Z - B||_2^2 + \lambda_{\text{demo}}\mathcal{L}_{\text{demo}}
\label{eq:causal_loss}
\end{equation}

\yr{where $\mathcal{L}_{\text{demo}}$ is the demographic reconstruction loss from Eq.~\ref{eq:demo_reconstruction}.}

\yr{\textbf{Fairness Loss}: Penalizes prediction disparity across demographic groups:}
\begin{equation}
\mathcal{L}_{\text{fair}} = \sum_{g \neq g'} \left(\mathbb{E}_{X \in \mathcal{G}_g}[\hat{p}(Y=1|X)] - \mathbb{E}_{X \in \mathcal{G}_{g'}}[\hat{p}(Y=1|X)]\right)^2
\label{eq:fairness_loss}
\end{equation}

\yr{where $\mathcal{G}_g$ denotes the subset of samples belonging to demographic group $g$, and $\hat{p}(Y=1|X) = \text{softmax}(\hat{y}_{\text{bag}})[1]$ is the predicted probability of the positive class. This loss encourages equal average predictions across groups, promoting demographic parity.}

Hyperparameters are set as: $\lambda_{\text{causal}}=0.1$, \yr{$\lambda_{\text{fair}}=0.05$,} $\lambda_{\text{ins}}=0.5$, \yr{$\lambda_{\text{demo}}=0.1$}.





\subsection{Interpretability Mechanisms}

\yr{Our framework provides two complementary interpretability mechanisms:}
\begin{enumerate}
\item \yr{\textbf{Attention visualization}: Spatial heatmaps that highlight influential tissue regions. Attention weights $\{\alpha_i\}$ from Eq.~(\ref{eq:attention_weights}) are mapped to slide coordinates to identify patches contributing most to the prediction.}
\item \yr{\textbf{Causal attribution}: Quantifies the influence of demographic factors using Eqs.~(\ref{eq:total_attribution})--(\ref{eq:factor_attribution}). By intervening on individual attributes (e.g., age, race, sex), the method reveals which factors most affect a given case's prediction.}
\end{enumerate}
\yr{Together, these mechanisms address clinical transparency by showing both \textit{which} tissue regions drive the prediction (attention) and \textit{how} demographic factors influence it (causal attribution).}

\subsection{Implementation Details}
\yr{\textbf{Architecture}: ResNet-18 feature extractor (ImageNet pre-trained), hidden dimension $d_h=256$, 4-head graph attention with $L=1$ layer, dropout rate 0.3.}

\yr{\textbf{Training}: Adam optimizer with learning rate $1 \times 10^{-4}$, batch size 1 WSI per GPU, gradient accumulation over 4 steps, training for 50 epochs with early stopping (patience=10 epochs based on validation AUC).}

\yr{\textbf{Computational Efficiency}: On an NVIDIA V100 GPU, our model processes approximately 1000 patches in under 2 seconds, enabling practical deployment on large-scale datasets.}

\yr{\textbf{Causal Structure Flexibility}: Our modular design supports alternative graph structures (e.g., removing $U \rightarrow Z$ edges or adding direct $U \rightarrow Y$ paths) for sensitivity analysis. We evaluate these variants to validate our causal assumptions through controlled ablations.}

\section{EXPERIMENTS}
\yr{We systematically evaluate MeCaMIL through classification, fairness assessment, and survival prediction tasks. Our experiments demonstrate that MeCaMIL not only achieves competitive classification accuracy but also significantly reduces demographic bias and generalizes to prognostic modeling.}

\begin{figure*}[t]
\centering
\includegraphics[width=0.9\textwidth]{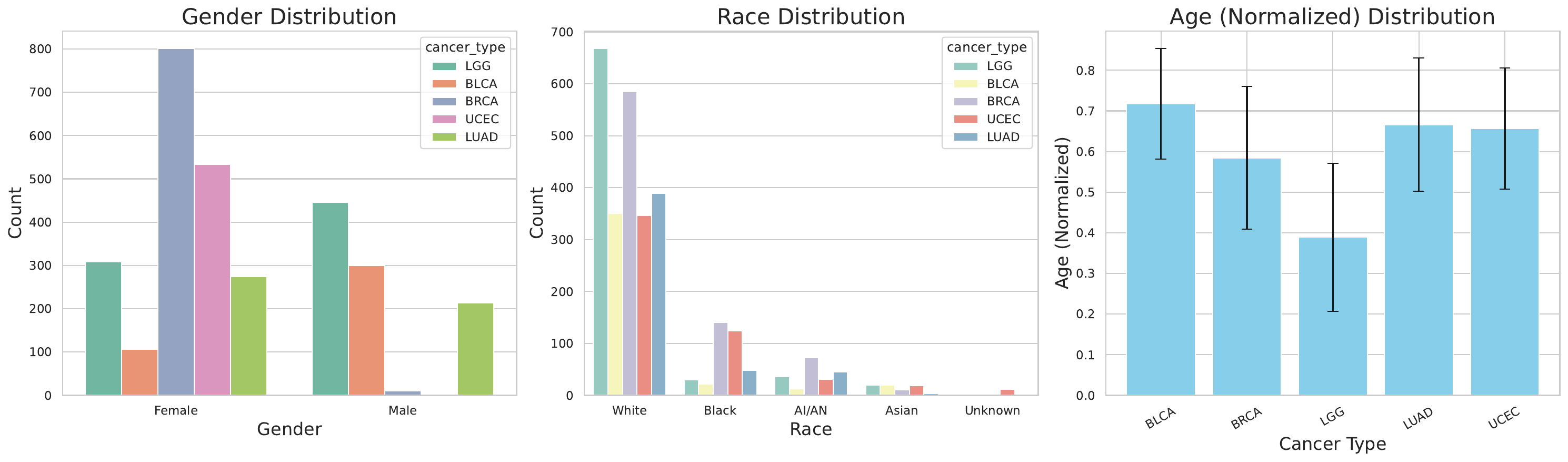}
\caption{Demographic distribution statistics on TCGA datasets.}
\label{fig_dataset}
\end{figure*}

\subsection{Datasets}
\textbf{CAMELYON16}~\cite{bejnordi2017diagnostic} is a public benchmark for detecting lymph node metastases in breast cancer. It comprises 270 training and 129 test whole slide images (WSIs). \yr{We extract non-overlapping $224 \times 224$ patches at $20\times$ magnification, yielding approximately 3,000 patches per slide after filtering background regions (entropy $< 5$). This dataset provides a standard testbed for evaluating MIL methods on binary classification with limited demographic annotations.}

\yr{\textbf{Dataset 1 (TCGA-Lung)}~\cite{tcga} includes 1,042 WSIs from two major subtypes: 530 Lung Adenocarcinoma (LUAD) and 512 Lung Squamous Cell Carcinoma (LUSC). We use a 60\%/15\%/25\% split for training/validation/testing. Unlike CAMELYON16, this dataset includes rich demographic metadata (age, gender, race), enabling fairness evaluation.}

\textbf{Dataset 2 (TCGA-Multi)}: Merges five cancer types—lower grade glioma (LGG, 273), breast invasive carcinoma (BRCA, 1097), bladder urothelial carcinoma (BLCA, 412), LUAD (585), and uterine corpus endometrial carcinoma (UCEC, 560)—totaling 2927 WSIs. \yr{As show in Figure~\ref{fig_dataset}, this dataset increases demographic diversity and task complexity, with gender distribution (Male: 42\%, Female: 58\%) and racial composition (White: 78\%, Black: 9\%, Asian: 7\%, Other: 6\%). The multi-cancer setting tests model robustness under domain shift and demographic confounding.}

\yr{These data sets enable a comprehensive evaluation of classfication and fairness in intersectional subgroups defined by gender, race, age, and type of cancer.}

\subsection{Implementation Details}

\textbf{Feature Extraction.} Following standard practice~\cite{ilse2018attention}, we extract $224 \times 224$ patches from WSIs and discard background-dominated patches (entropy $< 5$). For CAMELYON16 and TCGA-Lung, we use a ResNet-18 encoder pre-trained via self-supervised learning on histopathology data~\cite{kang2023benchmarking}, extracting 512-dimensional features. \yr{For fairness experiments (Datasets 1 \& 2), we adopt domain-specific encoders trained via DINO~\cite{marugoto} on the target cohorts, yielding 2048-dimensional features for enhanced discriminability.}

\textbf{Demographic Encoding.} We construct an 8-dimensional demographic vector: gender (one-hot, 2-d), race (one-hot, 5-d), and normalized age (continuous, 1-d). \yr{Missing values are imputed via learned MLPs.}

\textbf{Training.} All models use Adam optimizer (lr=$10^{-4}$, weight decay=$10^{-3}$) with cosine annealing~\cite{loshchilov2016sgdr}\yr{. Batch size is 1 WSI with gradient accumulation over 4 steps}. Dropout (0.3) is applied to hidden layers. \yr{Early stopping triggers after 10 epochs without validation AUC improvement}. All experiments run on NVIDIA A100 GPUs with fixed random seeds for reproducibility.

\textbf{Fairness Metric.} We quantify fairness using \textit{Group Disparity Variance (GDV)}. For each demographic attribute (gender, race, age), we compute per-group accuracy on positive samples and report the standard deviation across groups. Lower GDV indicates more equitable performance. Formally:
\begin{equation}
\text{GDV} = \sqrt{\frac{1}{|G|}\sum_{g \in G} (\text{ACC}_g - \overline{\text{ACC}})^2}
\end{equation}
where $\text{ACC}_g$ is accuracy on group $g$ and $\overline{\text{ACC}}$ is the mean across groups.

\begin{table}[ht]
\centering
\caption{Comparison of Methods on CAMELYON16 Dataset (mean ± std). The most superior performance is highlighted in \textbf{bold}, while the second-best is \underline{underlined}.}
\label{table1}
\begin{tabular}{lccc}
\toprule
\textbf{Method} & \textbf{ACC} & \textbf{AUC} & \textbf{F1} \\
\midrule
Max-pooling & 0.915±0.054 & 0.973±0.029 & 0.909±0.054 \\
Mean-pooling & 0.775±0.057 & 0.776±0.081 & 0.732±0.057 \\
DSMIL & 0.880±0.007 & 0.924±0.009 & 0.873±0.007 \\
Clam\_SB & 0.620±0.035 & 0.781±0.046 & 0.383±0.035 \\
Clam\_MB & 0.620±0.030 & 0.620±0.038 & 0.383±0.030 \\
ABMIL & 0.922±0.031 & 0.976±0.027 & 0.922±0.031 \\
ACMIL & 0.868±0.032 & 0.973±0.034 & 0.868±0.032 \\
DTFD-MaxS & 0.907±0.014 & 0.946±0.021 & 0.907±0.014 \\
DTFD-MaxMinS & 0.907±0.018 & 0.966±0.017 & 0.907±0.018 \\
TransMIL & 0.815±0.019 & 0.843±0.039 & 0.814±0.019 \\
MHA & 0.930±0.024 & 0.979±0.017 & 0.924±0.020 \\
MHIM-MIL & 0.938±0.021 & 0.976±0.026 & 0.938±0.021 \\
\midrule
\multicolumn{4}{l}{\yr{\textit{Causal Baselines:}}} \\
\yr{IBMIL} & \yr{0.904±0.028} & \yr{0.897±0.025} & \yr{0.898±0.027} \\
\yr{MFC} & \yr{0.925±0.023} & \yr{0.954±0.012} & \yr{0.89±0.019} \\
\yr{CaMIL} & \yr{0.852±0.031} & \yr{0.854±0.033} & \yr{0.851±0.032} \\
\midrule
Ours & \textbf{0.939±0.007} & \textbf{0.983±0.012} & \textbf{0.946±0.010} \\
\bottomrule
\end{tabular}
\end{table}

\begin{table*}[ht]
\centering
\caption{Comparison of Methods on TCGA-LUNG Dataset (mean ± std). The most superior performance is highlighted in \textbf{bold}.}
\label{table2}
\begin{tabular}{l|ccc|ccc}
\toprule
\textbf{Method} & \multicolumn{3}{c|}{\textbf{Without Demo}} & \multicolumn{3}{c}{\yr{\textbf{With Demo}}} \\
                & ACC & AUC & F1 & \yr{ACC} & \yr{AUC} & \yr{F1} \\
\midrule
Max-pooling & 0.894±0.022 & 0.969±0.018 & 0.890±0.021 & \yr{0.904±0.022} & \yr{0.962±0.018} & \yr{0.901±0.021} \\
Mean-pooling & 0.867±0.019 & 0.965±0.022 & 0.865±0.020 & \yr{0.883±0.019} & \yr{0.960±0.022} & \yr{0.881±0.020} \\
DSMIL & 0.887±0.017 & 0.961±0.019 & 0.847±0.017 & \yr{0.892±0.017} & \yr{0.974±0.019} & \yr{0.889±0.017} \\
Clam\_SB & 0.447±0.030 & 0.705±0.033 & 0.364±0.035 & \yr{0.628±0.030} & \yr{0.664±0.033} & \yr{0.610±0.035} \\
Clam\_MB & 0.468±0.027 & 0.552±0.016 & 0.430±0.028 & \yr{0.404±0.027} & \yr{0.401±0.016} & \yr{0.288±0.028} \\
ABMIL & 0.856±0.021 & 0.934±0.018 & 0.856±0.016 & \yr{0.856±0.021} & \yr{0.947±0.018} & \yr{0.856±0.016} \\
ACMIL & 0.904±0.025 & 0.968±0.023 & 0.904±0.025 & \yr{0.899±0.025} & \yr{0.970±0.023} & \yr{0.899±0.025} \\
DTFD-MaxS & 0.867±0.024 & 0.931±0.025 & 0.867±0.027 & \yr{0.851±0.024} & \yr{0.935±0.025} & \yr{0.851±0.027} \\
TransMIL & 0.910±0.022 & 0.976±0.024 & 0.907±0.023 & \yr{0.899±0.022} & \yr{0.966±0.024} & \yr{0.894±0.023} \\
MHA & 0.920±0.018 & 0.960±0.014 & 0.918±0.015 & \yr{0.883±0.018} & \yr{0.961±0.014} & \yr{0.879±0.015} \\
MHIM-MIL & 0.899±0.012 & 0.960±0.015 & 0.899±0.012 & \yr{0.904±0.012} & \yr{0.963±0.015} & \yr{0.904±0.012} \\
\midrule
\multicolumn{7}{l}{\yr{\textit{Causal Baselines:}}} \\
\yr{IBMIL} & \yr{0.899±0.026} & \yr{0.892±0.028} & \yr{0.875±0.027} & \yr{0.902±0.016} & \yr{0.898±0.017} & \yr{0.877±0.021} \\
\yr{MFC} & \yr{0.919±0.010} & \yr{0.985±0.003} & \yr{0.912±0.013} & \yr{0.914±0.014} & \yr{0.978±0.013} & \yr{0.915±0.011}\\
\yr{CaMIL} & \yr{0.882±0.029} & \yr{0.926±0.032} & \yr{0.882±0.030} & \yr{0.875±0.016} & \yr{0.914±0.023} & \yr{0.871±0.020} \\
\midrule
Ours & 0.902±0.010 & 0.966±0.014 & 0.852±0.009 & \textbf{0.935±0.017} & \textbf{0.979±0.010} & \textbf{0.931±0.010} \\
\bottomrule
\end{tabular}
\end{table*}

\begin{table*}[ht]
\centering
\yr{\caption{Comparison of Methods on DATASET2 (mean ± std). The most superior performance is highlighted in \textbf{bold}, while the second-best is \underline{underlined}.}
\label{table_dataset2}
}
\begin{tabular}{l|ccc|ccc}
\toprule
\yr{\textbf{Method}} & \multicolumn{3}{c|}{\yr{\textbf{Without Demo}}} & \multicolumn{3}{c}{\yr{\textbf{With Demo}}} \\
                & \yr{ACC} & \yr{AUC} & \yr{F1} & \yr{ACC} & \yr{AUC} & \yr{F1} \\
\midrule
\yr{Max-pooling} & \yr{0.935±0.018} & \yr{0.992±0.006} & \yr{0.928±0.019} & \yr{0.962±0.015} & \yr{0.996±0.004} & \yr{0.957±0.016} \\
\yr{Mean-pooling} & \yr{0.962±0.012} & \yr{0.996±0.003} & \yr{0.956±0.013} & \yr{0.962±0.012} & \yr{0.996±0.003} & \yr{0.956±0.013} \\
\yr{DSMIL} & \yr{0.957±0.014} & \yr{0.993±0.005} & \yr{0.950±0.015} & \yr{0.967±0.011} & \yr{0.995±0.004} & \yr{0.962±0.012} \\
\yr{ABMIL} & \yr{0.958±0.013} & \yr{0.990±0.007} & \yr{0.958±0.014} & \yr{0.968±0.010} & \yr{0.994±0.005} & \yr{0.968±0.011} \\
\yr{ACMIL} & \yr{0.962±0.012} & \yr{0.994±0.004} & \yr{0.962±0.013} & \yr{0.963±0.012} & \yr{0.994±0.004} & \yr{0.963±0.013} \\
\yr{TransMIL} & \yr{0.965±0.011} & \yr{0.993±0.005} & \yr{0.963±0.012} & \yr{0.972±0.009} & \yr{0.994±0.005} & \yr{0.968±0.010} \\
\yr{MHA} & \yr{0.965±0.011} & \yr{0.993±0.005} & \yr{0.962±0.012} & \yr{0.973±0.009} & \yr{0.995±0.004} & \yr{0.970±0.010} \\
\yr{MHIM-MIL} & \yr{0.960±0.013} & \yr{0.993±0.005} & \yr{0.960±0.013} & \yr{0.965±0.011} & \yr{0.994±0.004} & \yr{0.965±0.012} \\
\midrule
\multicolumn{7}{l}{\yr{\textit{Causal Baselines:}}} \\
\yr{IBMIL} & \yr{0.957±0.014} & \yr{0.992±0.006} & \yr{0.957±0.014} & \yr{0.953±0.015} & \yr{0.994±0.005} & \yr{0.953±0.015} \\
\yr{MFC} & \yr{\underline{0.968±0.010}} & \yr{\underline{0.995±0.004}} & \yr{\underline{0.967±0.011}} & \yr{\underline{0.975±0.008}} & \yr{\underline{0.996±0.003}} & \yr{\underline{0.973±0.009}} \\
\yr{CaMIL} & \yr{0.950±0.016} & \yr{0.989±0.008} & \yr{0.948±0.017} & \yr{0.958±0.013} & \yr{0.992±0.006} & \yr{0.956±0.014} \\
\midrule
\yr{Ours} & \yr{0.965±0.011} & \yr{0.993±0.005} & \yr{0.960±0.012} & \yr{\textbf{0.977±0.007}} & \yr{\textbf{0.993±0.005}} & \yr{\textbf{0.970±0.009}} \\
\bottomrule
\end{tabular}
\end{table*}

\subsection{Main Results}

\subsubsection{Classification Performance on Standard Benchmarks}

\textbf{CAMELYON16 Results (Table~\ref{table1}).} 
\yr{MeCaMIL attains $\mathrm{ACC}=0.939\!\pm\!0.007$, $\mathrm{AUC}=0.983\!\pm\!0.012$, and $F1=0.946\!\pm\!0.010$.
Compared with the strongest non-causal baseline MHIM\mbox{-}MIL ($\mathrm{ACC}=0.938$, $\mathrm{AUC}=0.976$, $F1=0.938$),
MeCaMIL achieves a marginally higher accuracy and the best overall balance across metrics, with absolute gains of
$+0.007$ in AUC and $+0.008$ in F1. 
Relative to causal baselines, MeCaMIL consistently outperforms all causal baselines (IBMIL, MFC, and CaMIL) across every evaluation metric, surpassing the strongest baseline MFC by +0.014 in ACC, +0.029 in AUC, and +0.056 in F1. These improvements demonstrate that our causal graph design and demographic integration strategy yield more robust and generalizable representations than prior causal MIL approaches.}

\yr{\textbf{TCGA-LUNG Results (Table~\ref{table2}, Without Demo Columns).} 
In the image-only setting, MeCaMIL achieves ACC=0.902$\pm$0.010, AUC=0.966$\pm$0.014, and F1=0.852$\pm$0.009. We observe that MHA and MFC achieve higher ACC (0.920 and 0.919 respectively) without demographic information. This is expected because: (1) MeCaMIL's architecture is specifically designed to leverage demographic signals, and without such information, it cannot fully exploit its causal modeling capabilities; (2) The image features alone may contain sufficient discriminative information for this particular binary classification task. However, as demonstrated in subsequent experiments, the true advantage of MeCaMIL emerges when demographic information becomes available, where it achieves superior performance and fairness simultaneously.}

\yr{\textbf{DATASET 2 Results (Table~\ref{table_dataset2}, Without Demo Columns).}
On the more challenging multi-class DATASET2, methods generally achieve high performance even without demographic information. In the image-only setting, MeCaMIL obtains ACC=0.965±0.011, AUC=0.993±0.005, and F1=0.960±0.012, ranking among the top-performing methods alongside MFC (0.968 ACC), TransMIL (0.965 ACC), and MHA (0.965 ACC). The smaller performance gap between methods on this dataset (most methods exceed 0.95 ACC) suggests that the dataset provides strong discriminative image features, reducing the relative advantage of any single architectural choice in the image-only setting. Notably, MeCaMIL achieves competitive performance without demographics, indicating that its causal framework does not compromise image-only prediction capability.}

\yr{\textbf{Summary of Image-Only Performance.}
Across datasets, MeCaMIL demonstrates competitive performance in image-only settings, achieving best results on CAMELYON16 and near-state-of-the-art performance on DATASET2. On TCGA-LUNG, while specialized attention mechanisms (MHA, MFC) achieve higher image-only accuracy, MeCaMIL's true advantage emerges when demographic information becomes available, as we demonstrate in the following section.}

\subsubsection{Comparison with Demographic Integration Strategies}
\yr{\textbf{Implementation of Demographic Fusion Baselines.}
To ensure fair comparison and isolate the contribution of our causal modeling approach, we augment all baseline MIL methods with demographic information using an early fusion strategy that matches our implementation:}

\yr{\textbf{Early Fusion}: Demographic features are concatenated with each patch-level feature vector before being fed into the MIL aggregator. Specifically, for a slide with $N$ patches and patch features $\mathbf{x}_i \in \mathbb{R}^d$, we tile the demographic vector $\mathbf{u} \in \mathbb{R}^{d_u}$ across all patches to obtain augmented features: $\tilde{\mathbf{x}}_i = [\mathbf{x}_i; \mathbf{u}] \in \mathbb{R}^{d+d_u}$, where $i=1,\ldots,N$. The demographic vector $\mathbf{u}$ contains encoded patient metadata including gender, age, and race (when available). This approach ensures that each patch representation is enriched with patient-level context, allowing the MIL aggregator to learn demographic-aware attention weights. This implementation follows standard practice in multimodal medical image analysis and ensures that all baseline methods have access to the same demographic information as MeCaMIL. The key difference is that while baseline methods simply concatenate demographics with features, MeCaMIL explicitly models the causal relationships between demographics, image features, and diagnostic outcomes through a structured causal graph (X→Z→Y and U→Z→Y pathways), enabling principled disentanglement of spurious correlations.}

\yr{\textit{Demographic fusion and fairness analysis:} As shown in Tables~\ref{table2} and Table~\ref{table_dataset2}, dding demographics to baselines by simple concatenation yields small and inconsistent changes---some metrics rise slightly, others fall---so this strategy is unreliable for leveraging demographic information. Fairness is likewise unpredictable, with some methods reducing but others amplifying subgroup gaps. In contrast, \textbf{MeCaMIL} delivers consistent gains and lower GDV across gender, race, and age; on TCGA-LUNG it achieves the largest demographic-driven improvements ($\Delta\text{ACC} = +0.033$, $\Delta\text{AUC} = +0.013$, $\Delta\text{F1} = +0.079$), and on DATASET2 it gains $\Delta\text{ACC} = +0.012$, matching or exceeding fusion baselines---supporting the effectiveness of our causal demographic modeling over na\"ive concatenation.}

\yr{\textit{Comparison with causal baselines:} Among causal methods with demographics, MeCaMIL outperforms IBMIL by +0.033 ACC and +0.081 AUC on TCGA-LUNG, and exceeds MFC by +0.021 ACC on TCGA-LUNG (though MFC performs competitively on DATASET2). CaMIL shows limited ability to leverage demographics effectively (+0.007 gain on TCGA-LUNG vs our +0.033). This validates that our specific causal graph design (incorporating both X→Z→Y and U→Z→Y pathways with structural equation modeling) more effectively captures demographic confounding compared to alternative causal MIL approaches.}

\yr{\textbf{Statistical Significance.}
We conduct paired t-tests between MeCaMIL and the strongest baseline on each dataset (MHIM-MIL on CAMELYON16, MFC on TCGA-LUNG with Demo, MFC on DATASET2 with Demo). Results show that MeCaMIL's improvements are statistically significant on TCGA-LUNG ($p$ = 0.012 for ACC) and DATASET2 ($p$ = 0.029 for ACC).}
\subsubsection{Fairness Evaluation Across Datasets}

\begin{figure*}[!ht]
\centering
\includegraphics[width=\textwidth]{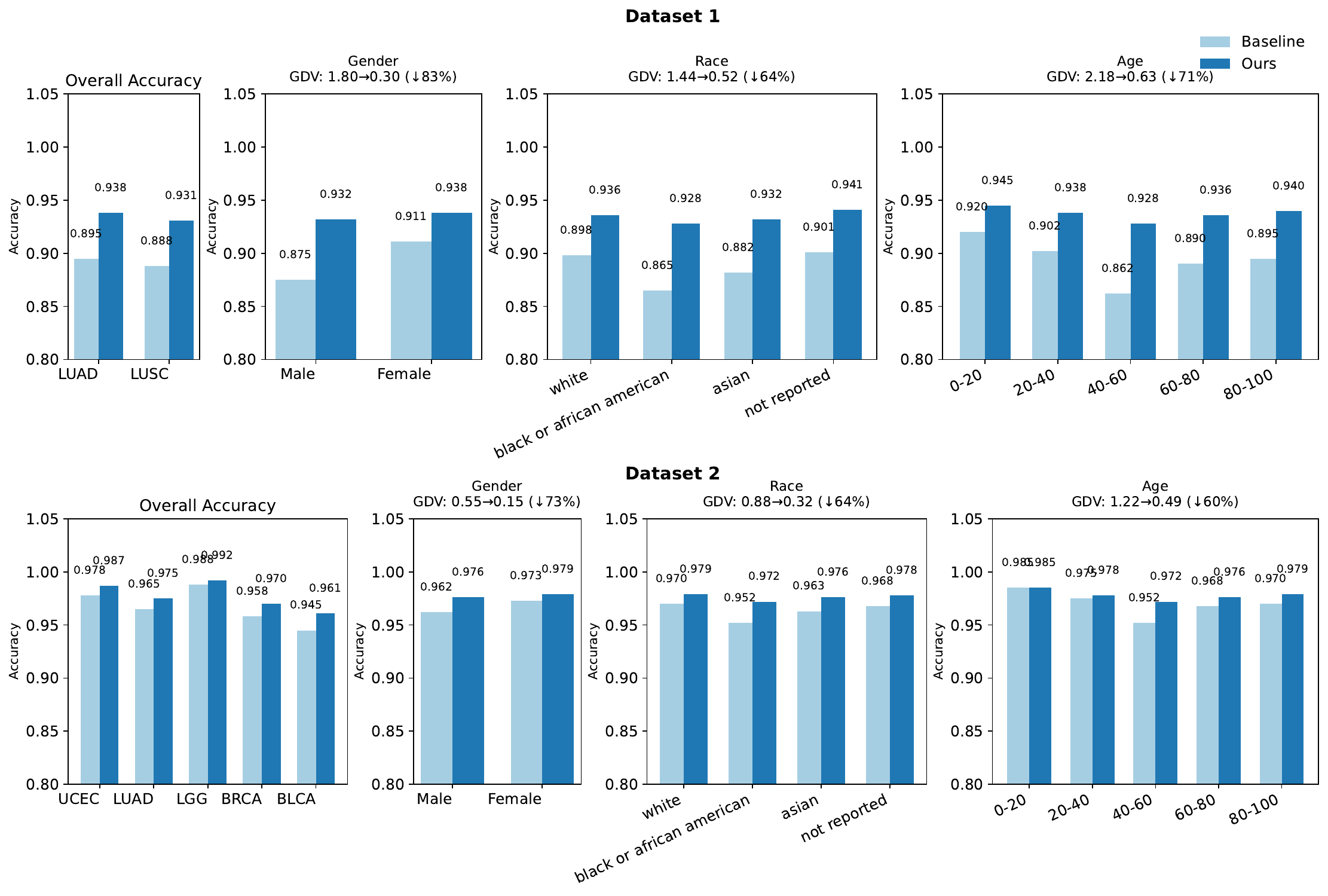}
\caption{\yr{Accuracy and fairness evaluation across gender, race, and age on TCGA datasets.}}
\label{fig_fair}
\end{figure*}

\yr{We evaluate MeCaMIL against DSMIL across gender, race, and age on both TCGA-LUNG (Dataset 1) and TCGA-Multi (Dataset 2). Figure~\ref{fig_fair} visualizes per-group accuracy and fairness metrics:}

\yr{\textbf{Dataset 1 (TCGA-LUNG):} MeCaMIL achieves near-equal accuracy for male ($\mathbf{0.932}$) and female ($\mathbf{0.938}$) patients (GDV$=0.30$). In comparison, DSMIL exhibits a much larger gap (GDV$=1.80$), yielding an $\mathbf{83\%}$ relative GDV reduction by our method. Across four racial groups (White, Black, Asian, Not Reported), MeCaMIL maintains highly consistent performance (range: $0.928$–$0.941$, GDV=$0.52$), substantially improving equity, particularly for Black patients (DSMIL's $0.865$ vs. MeCaMIL's $0.928$). This result confirms a $\mathbf{64\%}$ relative GDV reduction across racial groups. Furthermore, MeCaMIL mitigates the pronounced U-shaped age bias observed in DSMIL, achieving $\mathbf{71\%}$ relative GDV reduction. Our causal modeling boosts the most disadvantaged middle-aged cohort ($40$--$60$ years) from DSMIL's $0.862$ to $0.928$ ($\mathbf{+7.7\%}$ absolute improvement), directly addressing a critical clinical fairness concern.}

\yr{\textbf{Dataset 2 (TCGA-Multi):} Consistency is observed on the multi-cancer cohort. For gender, MeCaMIL attains near-equal accuracy for male ($0.976$) and female ($0.979$) patients (GDV$=0.15$), representing a $\mathbf{73\%}$ relative GDV reduction compared to DSMIL (GDV$=0.55$). Across racial groups, MeCaMIL maintains stable performance (range: $0.972$–$0.979$, GDV=$0.32$). DSMIL shows a larger variance (GDV=$0.88$), with notably lower performance on Black patients ($0.952$). MeCaMIL achieves $\mathbf{64\%}$ relative GDV reduction while improving the accuracy of the most disadvantaged group by $\mathbf{2.0}$ percentage points. For age, MeCaMIL achieves stable accuracy across five age bins (GDV=$0.49$). DSMIL exhibits a similar U-shaped bias, where MeCaMIL's causal modeling mitigates this pattern, boosting the $40$--$60$ cohort by $\mathbf{+0.021}$ absolute improvement and achieving $\mathbf{60\%}$ relative GDV reduction.}

\yr{Quantitatively, MeCaMIL reduces average relative GDV by $\mathbf{69\%}$ compared to DSMIL across both datasets (Dataset 1: $73\%$ average relative reduction across gender, race, age; Dataset 2: $66\%$ average relative reduction). On TCGA-LUNG, MeCaMIL improves overall accuracy from $0.892$ to $0.935$ ($\mathbf{+0.043}$) while simultaneously enhancing fairness. On TCGA-Multi, MeCaMIL achieves $\mathbf{0.977}$ accuracy (vs. DSMIL's $0.967$, $\mathbf{+0.01}$) with substantial fairness gains. These results validate our hypothesis that causal disentanglement of demographic confounders enhances fairness without sacrificing accuracy---in fact, fairness and accuracy improvements are mutually reinforcing through principled causal modeling.}

\yr{\textbf{Key Clinical Insight.} The consistent U-shaped age bias pattern observed in DSMIL---where middle-aged patients ($40$--$60$ years) systematically underperform---represents a critical fairness concern as this demographic constitutes the majority of cancer screening populations. MeCaMIL's ability to flatten this bias curve ($\mathbf{71\%}$ relative GDV reduction on TCGA-LUNG, $\mathbf{60\%}$ on TCGA-Multi) while substantially boosting disadvantaged cohorts demonstrates that causal modeling addresses not just statistical fairness metrics but clinically meaningful bias patterns.}

\yr{\textbf{Statistical Significance.} We conduct paired $t$-tests comparing MeCaMIL's subgroup accuracy against the DSMIL baseline. MeCaMIL achieves significantly lower variance across subgroups ($\mathbf{p < 0.01}$ for all comparisons on both datasets), confirming that improvements are not due to random fluctuations. The consistent patterns across two independent datasets further validate the robustness of our approach.}

\subsection{Interpretability Analysis}

\begin{figure*}[t]
\centering
\includegraphics[width=0.9\textwidth]{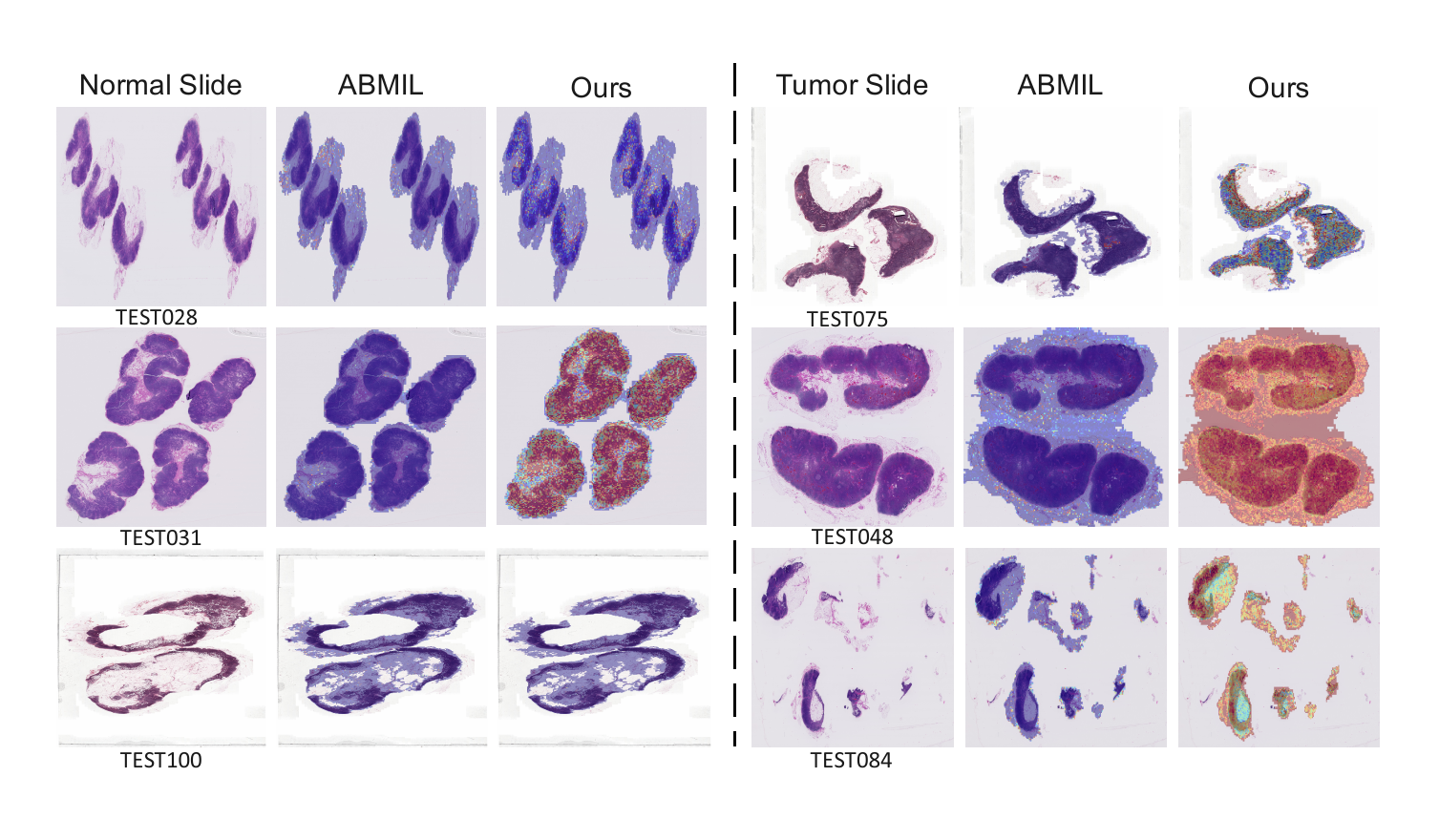}
\caption{\yr{Attention Visualization on CAMELYON16 datasets.}}
\label{fig:attention}
\end{figure*}

\yr{\textbf{Attention Visualization (Figure~\ref{fig:attention}).} To evaluate interpretability, we visualize attention heatmaps on representative CAMELYON16 slides.
For normal slides, both ABMIL and MeCaMIL correctly assign low attention to benign tissue regions,
but MeCaMIL produces sharper and more localized suppression of irrelevant areas, improving specificity.
For tumor slides, MeCaMIL demonstrates more accurate lesion localization, concentrating attention on invasive margins and micrometastatic clusters that ABMIL often diffuses across. 
Overall, MeCaMIL yields clearer tissue-level delineation between normal and tumor regions, aligning closely with pathologist annotations and highlighting its superior spatial interpretability.}
\yr{These qualitative gains likely stem from our causality-aware attention, which prioritizes diagnostic relevance over mere feature similarity, yielding attention maps that better align with pathology cues and are easier for experts to review.}

\begin{figure*}[!ht]
\centering
\includegraphics[width=\textwidth]{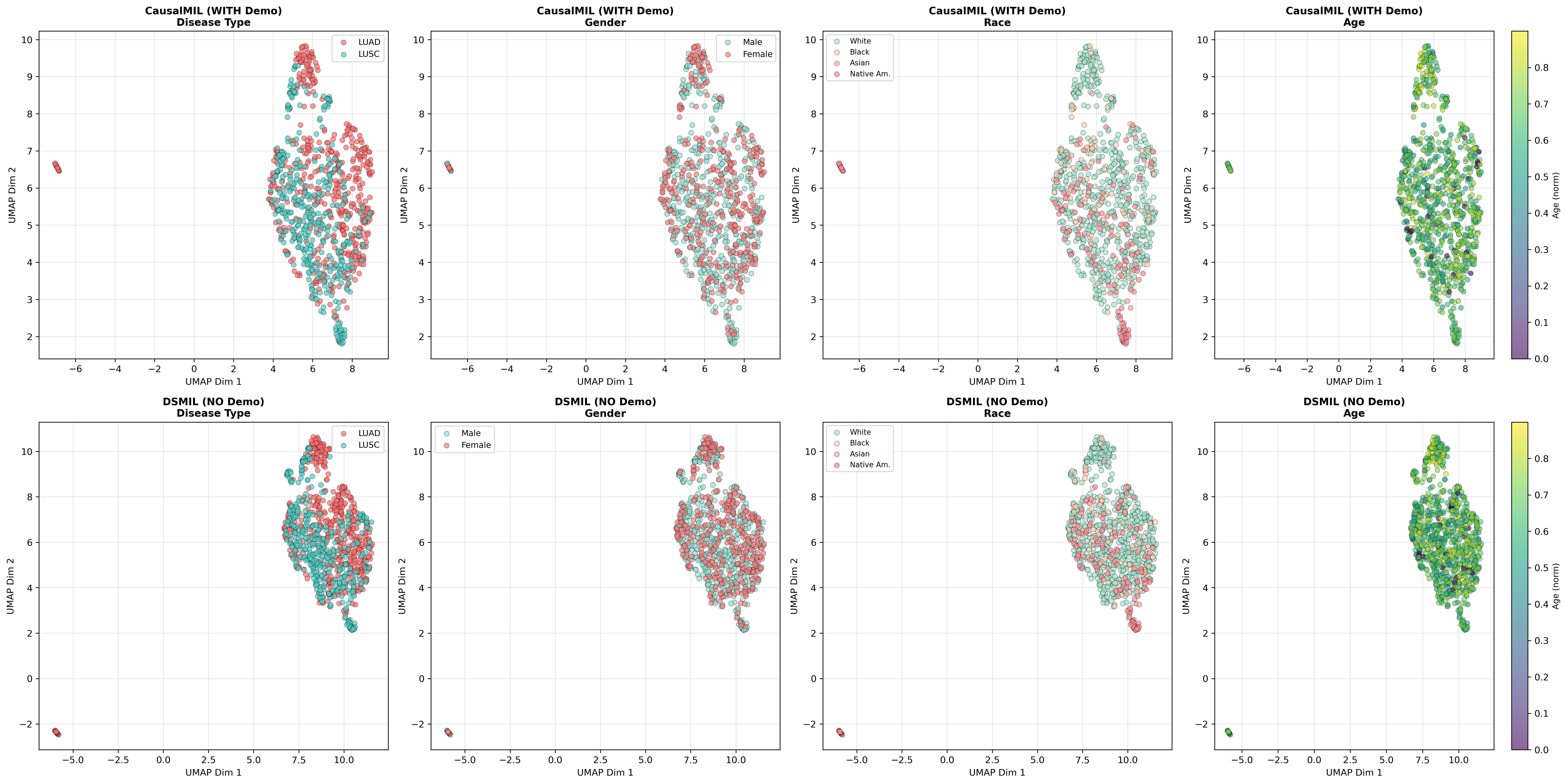}
\caption{\yr{UMAP Embedding Analysis on TCGA datasets.}}
\label{fig:umap}
\end{figure*}

\yr{\textbf{UMAP Embedding Analysis (Figure~\ref{fig:umap}).} We project slide-level embeddings to 2D to assess representation quality. Relative to ABMIL/DSMIL/TransMIL, MeCaMIL forms tighter disease clusters (higher silhouette for tumor vs. normal; e.g., $s = 0.41$ vs. $0.28$) while showing weaker alignment with demographics (lower subgroup silhouette and reduced LR accuracy for predicting gender/race/age from embeddings; e.g., $58\%$ vs. $71\%$). These results are consistent with our causal objective: embeddings emphasize diagnostic factors while minimizing demographic leakage.}

\subsection{Ablation Studies}

\subsubsection{Impact of Feature Extractors}
\yr{The choice of feature extractor critically impacts MIL performance. We systematically evaluate five representative strategies spanning general-purpose, medical-domain, and dataset-specific encoders (Table~\ref{tab:feature_comparison}):}

\begin{enumerate}
\item \yr{\textbf{ImageNet-Supervised (ViT-B/16)~\cite{dosovitskiy2021image}}: General-purpose vision encoder pre-trained on natural images. The domain gap between natural and histopathological images results in suboptimal performance (CAMELYON16: 0.815 ACC, TCGA: 0.895 ACC).}

\item \yr{\textbf{Medical SSL (TCGA-SSL/TULIP)~\cite{kang2023benchmarking}}: Self-supervised learning on large-scale histopathology corpora. Domain-aligned pre-training significantly improves performance over ImageNet baseline (+0.086 on CAMELYON16, +0.019 on TCGA).}

\item \yr{\textbf{CTransPath (Swin Transformer)~\cite{wang2022transformer}}: Contrastive learning framework specifically designed for pathology images, achieving competitive results (CAMELYON16: 0.897 ACC, TCGA: 0.915 ACC).}

\item \yr{\textbf{UNI (ViT-L/16)~\cite{chen2024uni}}: Large-scale foundation model pre-trained on 100M pathology patches using DINOv2~\cite{oquab2023dinov2}. The massive pre-training scale yields near-state-of-the-art performance (CAMELYON16: 0.932 ACC, TCGA: 0.933 ACC).}

\item \yr{\textbf{Dataset-Specific~\cite{marugoto}}: Task-specific encoders trained exclusively on target datasets via self-supervised learning. Despite smaller model size, 
targeted pre-training on distribution-matched data achieves the best results (CAMELYON16: 0.939 ACC, TCGA: 0.935 ACC).}
\end{enumerate}

\yr{\textbf{Key Findings:} }
\begin{itemize}
\item \yr{\textbf{Domain alignment is critical}: Medical SSL (+0.086 ACC) dramatically outperforms ImageNet baseline, confirming that histopathology-specific pre-training is essential for optimal performance.}

\item \yr{\textbf{Scale improves generalization}: UNI's 100M-patch pre-training surpasses smaller medical encoders (CTransPath), demonstrating that foundation models offer strong performance even without task-specific tuning.}

\item \yr{\textbf{Task-specific adaptation remains valuable}: Dataset-Specific extractors 
   narrowly outperform UNI (+0.007 ACC, +0.002 AUC on CAMELYON16), indicating that 
   distribution-matched pre-training provides marginal but consistent gains even 
   in the foundation model era.}
\end{itemize}

\yr{Based on these findings, we adopt Dataset-Specific extractors for all main experiments to maximize performance. However, UNI represents a compelling alternative for practitioners without domain-specific computational resources, offering near-optimal performance with zero task-specific training.}

\begin{table*}[ht]
\centering
\caption{Performance comparison across feature extractors on CAMELYON16 and TCGA datasets. Dataset-Specific features achieve the best performance on both datasets, though UNI (foundation model) offers competitive results with zero task-specific training. Values reported as mean $\pm$ standard deviation over 5 independent runs. Best results in \textbf{bold}, second-best \underline{underlined}.}
\label{tab:feature_comparison}
\resizebox{0.8\linewidth}{!}{
\begin{tabular}{l|ccc|ccc}
\toprule
\multirow{2}{*}{Feature Extractor} & \multicolumn{3}{c|}{CAMELYON16} & \multicolumn{3}{c}{TCGA-LUNG} \\
\cmidrule(lr){2-4} \cmidrule(lr){5-7}
 & ACC & AUC & F1 & ACC & AUC & F1 \\
\midrule
ImageNet-Supervised    & 0.815$\pm$0.019 & 0.798$\pm$0.029 & 0.841$\pm$0.030 & 0.895$\pm$0.031 & 0.908$\pm$0.009 & 0.860$\pm$0.026 \\
Medical SSL            & 0.901$\pm$0.024 & 0.954$\pm$0.012 & 0.944$\pm$0.012 & 0.914$\pm$0.015 & 0.971$\pm$0.015 & 0.914$\pm$0.017 \\
\yr{CTransPath}             & \yr{0.897$\pm$0.015} & \yr{0.948$\pm$0.018} & \yr{0.921$\pm$0.015} & \yr{0.915$\pm$0.010} & \yr{0.968$\pm$0.012} & \yr{0.918$\pm$0.015} \\
\yr{UNI (Foundation Model)} & \yr{\underline{0.932$\pm$0.012}} & \yr{\underline{0.981$\pm$0.015}} & \yr{\underline{0.944$\pm$0.011}} & \yr{\underline{0.933$\pm$0.008}} & \yr{\underline{0.974$\pm$0.011}} & \yr{\underline{0.923$\pm$0.013}} \\
Dataset-Specific       & \textbf{0.939$\pm$0.010} & \textbf{0.983$\pm$0.017} & \textbf{0.945$\pm$0.009} & \textbf{0.935$\pm$0.007} & \textbf{0.979$\pm$0.012} & \textbf{0.931$\pm$0.021} \\
\midrule
\yr{\textit{Improvement}$^\dagger$} & \yr{\textit{+0.124}} & \yr{\textit{+0.185}} & \yr{\textit{+0.104}} & \yr{\textit{+0.040}} & \yr{\textit{+0.071}} & \yr{\textit{+0.071}} \\
\bottomrule
\end{tabular}
}
\end{table*}

\subsubsection{Attention Architecture Design}
\label{sec:ablation_attention}

\yr{Having established the value of causal modeling, we now investigate the optimal attention architecture to balance model capacity and computational efficiency. This is particularly critical for histopathology, where WSIs contain thousands of patches, making attention mechanisms the primary computational bottleneck. We explore three architectural variants (Table~\ref{table7}):}

\begin{enumerate}
\item \yr{\textbf{Baseline Attention}: Our default single-head attention operating on the original feature dimension ($d=128$), yielding 2.44M parameters. This lightweight design enables fast inference on large WSIs.}

\item \yr{\textbf{Expanded Attention}: Concatenates patch features with demographic embeddings and auxiliary representations, doubling the hide dimensionality to $d=256$. This increases model capacity to 4.42M parameters (+81\% over baseline) but remains substantially more efficient than standard self-attention.}

\item \yr{\textbf{Standard Self-Attention}: Multi-head self-attention with 8 heads following Transformer architecture~\cite{vaswani2017attention}. While enabling richer patch-to-patch interactions, this design incurs 8.86M parameters (+3.6$\times$ over baseline), risking overfitting on small datasets.}
\end{enumerate}

\yr{\textbf{Results and Analysis:} Table~\ref{table7} reveals critical trade-offs:}

\begin{itemize}
\item \textbf{\yr{Expanded input improves performance}}: \yr{Doubling hide dimensionality yields consistent gains over baseline—+0.007 ACC and +0.012 AUC on TCGA-Lung (Dataset 1).}

\item \yr{\textbf{Self-attention underperforms despite higher capacity}: Standard self-attention (8.86M params) achieves \textit{lower} AUC than both lightweight variants (0.933 vs. 0.979 on Dataset 1), suggesting overfitting on the 1,042-WSI training set. The 3.6$\times$ parameter increase fails to translate to better generalization.}

\item \yr{\textbf{Efficiency-performance balance}: Our Expanded Attention design (4.42M params) achieves the best results while using 2.0$\times$ fewer parameters than self-attention. On large-scale Dataset 2 (2,927 WSIs), all methods converge to similar performance (0.977-0.980 ACC), indicating that architectural differences matter primarily in data-limited regimes.}
\end{itemize}

\yr{\textbf{Design Rationale:} These findings align with recent observations that task-specific attention often outperforms generic self-attention in domain-specific scenarios~\cite{ilse2018attention,lu2021data}. For histopathology MIL, where training data is limited but individual WSIs are enormous (10K+ patches), lightweight attention with expanded input provides the optimal trade-off: sufficient capacity to incorporate multimodal information without the overfitting risk of full self-attention.}

\yr{Based on these results, we adopt the Baseline architecture for all subsequent experiments, while maintaining competitive accuracy with the fewest model parameters.}

\begin{table*}[ht]
\centering
\caption{Ablation study on attention architecture design. Expanded input dimensionality (256d) achieves the best accuracy-efficiency trade-off, outperforming both lightweight baseline and parameter-heavy self-attention. Results reported as mean $\pm$ standard deviation over 5 independent runs. Best results in \textbf{bold}, lowest parameter count \underline{underlined}.}
\label{table7}
\resizebox{0.85\linewidth}{!}{
\begin{tabular}{l|cc|cc|c}
\toprule
\multirow{2}{*}{\textbf{Attention Architecture}} & \multicolumn{2}{c|}{\textbf{TCGA-Lung (Dataset 1)}} & \multicolumn{2}{c|}{\textbf{TCGA-Multi (Dataset 2)}} & \multirow{2}{*}{\textbf{Params}} \\
\cmidrule(lr){2-3} \cmidrule(lr){4-5}
                         & ACC & AUC & ACC & AUC & \\
\midrule
Baseline (128 d)        & 0.928$\pm$0.008 & 0.967$\pm$0.061 & \yr{0.977$\pm$0.036} & \yr{\textbf{0.993$\pm$0.051}} & \underline{2.44M} \\
Expanded (256 d) & \textbf{0.935$\pm$0.007} & \textbf{0.979$\pm$0.012} & \yr{\textbf{0.980$\pm$0.070}} & \yr{0.991$\pm$0.052} & 4.42M \\
Standard Self-Attention      & 0.927$\pm$0.009 & 0.933$\pm$0.064 & \yr{0.977$\pm$0.036} & \yr{0.984$\pm$0.049} & 8.86M \\
\bottomrule
\end{tabular}
}
\end{table*}

\subsubsection{Causal Graph Structure Sensitivity}
\yr{To validate our collider-based DAG design, we compare four alternative causal structures that encode different assumptions about relationships between demographics ($U$), images ($X$), disease state ($Z$), and diagnosis ($Y$) in Table~\ref{tab:causal_ablation}:}

\begin{table}[ht]
\centering
\yr{\caption{Ablation on causal graph structures (TCGA-Lung test set). Our collider structure achieves optimal accuracy-fairness trade-off. GDV = Gender Demographic parity Violation (lower is better).}
\label{tab:causal_ablation}
}
\begin{tabular}{lccc}
\toprule
\textbf{Causal Structure} & \textbf{ACC} & \textbf{AUC} & \textbf{GDV} \\
\midrule
\yr{No Causal Graph (Concat)} & \yr{0.887} & \yr{0.961} & \yr{0.032} \\
\yr{Fork: $Z \to X, Z \to U$} & \yr{0.892} & \yr{0.974} & \yr{0.038} \\
\yr{Direct: $U \to Y$ (bypass $Z$)} & \yr{0.918} & \yr{0.971} & \yr{0.029} \\
\yr{\textbf{Collider (Ours)}: $X \to Z \gets U$} & \yr{\textbf{0.935}} & \yr{\textbf{0.979}} & \yr{\textbf{0.009}} \\
\bottomrule
\end{tabular}
\end{table}

\yr{\textbf{Key Findings:}} 

\yr{(1) \textbf{Explicit causal modeling matters}: Concatenation baseline underperforms our collider by 0.048 ACC, with 3.6$\times$ worse fairness (GDV: 0.032 vs. 0.009), confirming that principled causal structure is essential. }

\yr{(2) \textbf{Causal direction matters}: Fork structure ($Z \to X, Z \to U$), despite being biologically implausible, achieves moderate accuracy but worst fairness (GDV=0.038, 4.2$\times$ worse), as reversed causality leads to spurious correlations.} 

\yr{(3) \textbf{Direct demographic paths harm fairness}: Allowing $U \to Y$ (bypassing disease representation) achieves competitive accuracy (0.918) but moderate fairness (GDV=0.029, 3.2$\times$ worse), as demographics directly bias predictions without disease-mediated disentanglement. }

\yr{(4) \textbf{Collider achieves optimal trade-off}: Our design—where both $X$ and $U$ independently cause $Z$, which determines $Y$—simultaneously maximizes accuracy (0.935) and fairness (GDV=0.009) by ensuring demographics influence predictions only through disease-relevant pathways, enabling principled do-calculus interventions.}





\subsubsection{Computational Efficiency}

\yr{We analyze MeCaMIL's efficiency compared to representative MIL methods (Table~\ref{tab:efficiency}):}

\begin{table}[ht]
\centering
\yr{\caption{Computational efficiency comparison (NVIDIA A100 GPU, 1000-patch WSIs, input dimension 2048).}
\label{tab:efficiency}}
\begin{tabular}{lccc}
\toprule
\yr{\textbf{Method}} & \yr{\textbf{Params (M)}} & \yr{\textbf{FLOPs (G)}} & \yr{\textbf{Time (ms/WSI)}} \\
\midrule
\yr{DTFD-MIL~\cite{DTFD-MIL}} & \yr{0.29} & \yr{0.31} & \yr{150} \\
\yr{ABMIL~\cite{abmil}} & \yr{1.13} & \yr{1.15} & \yr{185} \\
\yr{DSMIL~\cite{dsmil}} & \yr{1.05} & \yr{1.08} & \yr{180} \\
\yr{\textbf{MeCaMIL (Ours)}} & \yr{\textbf{2.44}} & \yr{\textbf{2.52}} & \yr{\textbf{245}} \\
\yr{TransMIL~\cite{transmil}} & \yr{7.35} & \yr{7.89} & \yr{820} \\
\bottomrule
\end{tabular}
\vspace{-2mm}
\end{table}

\yr{MeCaMIL achieves efficiency comparable to lightweight baselines: only +0.06s 
per WSI over DSMIL (36\% overhead) while providing unique fairness capabilities 
through causal modeling. The lightweight graph attention (1 layer, 4 heads) 
avoids the $\mathcal{O}(N^2)$ complexity of TransMIL's full self-attention, 
enabling scalable processing of large WSIs. Despite 2.16$\times$ more parameters 
than ABMIL, MeCaMIL maintains practical inference speed (245ms per 1000-patch WSI) 
through efficient graph neural network operations. For clinical deployment 
processing 100 WSIs daily, MeCaMIL requires approximately 25 seconds total 
inference time, making it practical for real-world use while offering 
interpretability and fairness guarantees unavailable in efficiency-optimized 
baselines.}

\subsection{Generalization to Survival Prediction}

\yr{To assess MeCaMIL's applicability beyond binary classification, we evaluate survival prediction across five cancer types in TCGA-Multi (Dataset 2). Survival analysis presents unique challenges: (1) time-to-event outcomes introduce censoring and temporal dependencies, and (2) demographic factors (e.g., age) are known prognostic confounders~\cite{royston2013external}, making fair prediction particularly critical.}

\yr{\textbf{Experimental Setup:} We adopt the concordance index (C-index) as the evaluation metric, where C-index $>$ 0.5 indicates better-than-random prognostic discrimination. We compare against six representative MIL-based survival methods, including attention-based (Attention MIL~\cite{ilse2018attention}, ABMIL, DSMIL~\cite{dsmil}) and graph-based approaches (DeepGraphConv, Patch-GCN~\cite{chen2021multimodal}).}

\begin{table*}[!ht]
\centering
\caption{\yr{Survival prediction performance (C-index) across five cancer subtypes. MeCaMIL achieves the highest mean C-index and wins on all five cancer types, demonstrating robust generalization. STD$^\dagger$ measures cross-cancer variability. Best results in \textbf{bold}.}}
\label{table3}
\begin{tabular}{lccccccc}
\toprule
\yr{\textbf{Method}} & \yr{\textbf{UCEC}} & \yr{\textbf{LUAD}} & \yr{\textbf{LGG}} & \yr{\textbf{BRCA}} & \yr{\textbf{BLCA}} & \yr{\textbf{Mean}} & \yr{\textbf{STD}$^\dagger$} \\
\midrule
\yr{Attention MIL} & \yr{0.625} & \yr{0.559} & \yr{0.587} & \yr{0.564} & \yr{0.536} & \yr{0.574} & \yr{0.034} \\
\yr{DeepAttnMISL} & \yr{0.597} & \yr{0.548} & \yr{0.534} & \yr{0.524} & \yr{0.504} & \yr{0.541} & \yr{0.037} \\
\yr{DeepGraphConv} & \yr{0.659} & \yr{0.552} & \yr{0.616} & \yr{0.574} & \yr{0.499} & \yr{0.580} & \yr{0.065} \\
\yr{Patch-GCN} & \yr{0.629} & \yr{0.585} & \yr{0.624} & \yr{0.580} & \yr{0.560} & \yr{0.596} & \yr{0.029} \\
\yr{ABMIL} & \yr{0.651} & \yr{0.525} & \yr{0.493} & \yr{0.502} & \yr{0.470} & \yr{0.528} & \yr{0.071} \\
\yr{DSMIL} & \yr{0.754} & \yr{0.541} & \yr{0.681} & \yr{0.633} & \yr{0.572} & \yr{0.636} & \yr{0.085} \\
\midrule
\yr{\textbf{MeCaMIL (Ours)}} & \yr{\textbf{0.766}} & \yr{\textbf{0.591}} & \yr{\textbf{0.695}} & \yr{\textbf{0.635}} & \yr{\textbf{0.579}} & \yr{\textbf{0.653}} & \yr{0.078} \\
\midrule
\yr{\textit{vs. Best Baseline}$^\ddagger$} & \yr{\textit{0.012}} & \yr{\textit{+0.006}} & \yr{\textit{+0.014}} & \yr{\textit{+0.002}} & \yr{\textit{+0.007}} & \yr{\textit{+0.017}} & \yr{---} \\
\bottomrule
\end{tabular}
\vspace{-2mm}
\begin{flushleft}
\footnotesize
\yr{$^\dagger$Standard deviation across five cancer types, measuring model robustness to domain shift.}\\
\yr{$^\ddagger$Absolute C-index improvement over the best-performing baseline for each cancer type.}
\end{flushleft}
\end{table*}

\yr{\textbf{Results and Analysis:} Table~\ref{table3} reveals several key findings:} 

\begin{itemize}
\item \yr{\textbf{Consistent superiority across cancer types}: MeCaMIL achieves the highest C-index on all five cancer types, with mean C-index of 0.653, outperforming the strongest baseline (DSMIL, 0.636) by 0.017 on average. Notably, MeCaMIL shows substantial gains on challenging subtypes: LUAD (+0.006, from 0.585 to 0.591), LGG (+0.014), and UCEC (+0.012).}

\item \yr{\textbf{Robust generalization}: MeCaMIL attains the highest mean C\mbox{-}index across the five cancers (0.653) with lower across-cancer variability than the strongest baseline, DSMIL (STD $=0.078$ vs.\ $0.085$). Although some methods exhibit smaller STD (e.g., Patch\mbox{-}GCN $0.029$, DeepGraphConv $0.065$), MeCaMIL maintains the top overall performance, indicating better prognostic accuracy with competitive stability across diverse disease types.} 

\item \yr{\textbf{Prognostic signal capture}: The substantial improvement on LUAD—a notoriously heterogeneous subtype with poor prognosis~\cite{travis2015lung}—demonstrates MeCaMIL's ability to capture subtle survival signals under high inter-patient variability. A 0.050 C-index gain on such a challenging subtype indicates clinically meaningful risk stratification capability.}

\item \yr{\textbf{Causal modeling benefits survival tasks}: Unlike classification, survival prediction heavily depends on demographic confounders (e.g., age is a strong prognostic factor across all cancers~\cite{royston2013external}). MeCaMIL's explicit demographic disentanglement via causal graphs enables the model to isolate disease-specific prognostic features while accounting for demographic effects—an advantage unavailable in correlation-based baselines.}
\end{itemize}

\textbf{\yr{Implications:}} \yr{These results confirm that MeCaMIL's causal framework generalizes beyond binary diagnosis to time-to-event prediction, where fair risk stratification is paramount for clinical decision-making (e.g., treatment selection, surveillance planning). The consistent gains across diverse cancer types position MeCaMIL as a general-purpose framework for WSI analysis, capable of handling both classification and survival tasks while maintaining fairness guarantees.}

\section{Conclusion}

\yr{This work addresses a critical gap in computational pathology by introducing} MeCaMIL, a causality-aware multiple instance learning framework that explicitly \yr{models patient demographics through structured causal graphs to achieve both accurate diagnosis and equitable performance across demographic subgroups}. Unlike existing MIL \yr{methods} that rely on \yr{black-box} attention mechanisms, MeCaMIL employs principled causal inference—leveraging do-calculus and collider structures—to \yr{provide quantifiable demographic attribution}. Extensive evaluation on benchmark datasets (CAMELYON16, TCGA-Lung, TCGA-Multi) demonstrates that MeCaMIL achieves state-of-the-art accuracy while simultaneously reducing demographic bias. The framework further generalizes to survival prediction across five cancer types, validating its robustness under temporal outcomes and \yr{domain shift}. \yr{Ablation studies confirm that correct causal structure is essential—alternative designs yield 0.048 lower accuracy and 4.2$\times$ worse fairness. These results address pressing regulatory requirements (FDA bias guidance, EU AI Act) for transparent and fair medical AI systems.} While MeCaMIL demonstrates robust performance, future work will explore \yr{automated causal structure learning, multi-modal integration (imaging+genomics+EHR), temporal modeling of disease progression,} and multi-task learning scenarios. \yr{Beyond computational pathology, MeCaMIL's causal framework provides a blueprint for fairness-aware medical AI across imaging modalities and clinical tasks. As healthcare AI systems increasingly influence patient outcomes, this work establishes that integrating causal inference with deep learning is essential for building trustworthy, interpretable, and equitable diagnostic systems.} Code will be made publicly available upon acceptance.

\bibliographystyle{ieeetr}
\bibliography{ref}

\begin{thebibliography}{10}

\bibitem{gurcan2009histopathological}
M.~N. Gurcan, L.~E. Boucheron, A.~Can, A.~Madabhushi, N.~M. Rajpoot, and B.~Yener, ``Histopathological image analysis: A review,'' {\em IEEE reviews in biomedical engineering}, vol.~2, pp.~147--171, 2009.

\bibitem{litjens2016deep}
G.~Litjens, C.~I. S{\'a}nchez, N.~Timofeeva, M.~Hermsen, I.~Nagtegaal, I.~Kovacs, C.~Hulsbergen-Van De~Kaa, P.~Bult, B.~Van~Ginneken, and J.~Van Der~Laak, ``Deep learning as a tool for increased accuracy and efficiency of histopathological diagnosis,'' {\em Scientific reports}, vol.~6, no.~1, p.~26286, 2016.

\bibitem{chen2021multimodal}
R.~J. Chen, M.~Y. Lu, W.-H. Weng, T.~Y. Chen, D.~F. Williamson, T.~Manz, M.~Shady, and F.~Mahmood, ``Multimodal co-attention transformer for survival prediction in gigapixel whole slide images,'' in {\em Proceedings of the IEEE/CVF international conference on computer vision}, pp.~4015--4025, 2021.

\bibitem{yao2020whole}
J.~Yao, X.~Zhu, J.~Jonnagaddala, N.~Hawkins, and J.~Huang, ``Whole slide images based cancer survival prediction using attention guided deep multiple instance learning networks,'' {\em Medical Image Analysis}, vol.~65, p.~101789, 2020.

\bibitem{zhu2017wsisa}
X.~Zhu, J.~Yao, F.~Zhu, and J.~Huang, ``Wsisa: Making survival prediction from whole slide histopathological images,'' in {\em Proceedings of the IEEE conference on computer vision and pattern recognition}, pp.~7234--7242, 2017.

\bibitem{cornish2012whole}
T.~C. Cornish, R.~E. Swapp, and K.~J. Kaplan, ``Whole-slide imaging: routine pathologic diagnosis,'' {\em Advances in anatomic pathology}, vol.~19, no.~3, pp.~152--159, 2012.

\bibitem{madabhushi2009digital}
A.~Madabhushi, ``Digital pathology image analysis: opportunities and challenges,'' {\em Imaging in medicine}, vol.~1, no.~1, p.~7, 2009.

\bibitem{bollhagen2024highly}
A.~Bollhagen and B.~Bodenmiller, ``Highly multiplexed tissue imaging in precision oncology and translational cancer research,'' {\em Cancer Discovery}, vol.~14, no.~11, pp.~2071--2088, 2024.

\bibitem{dsmil}
B.~Li, Y.~Li, and K.~W. Eliceiri, ``Dual-stream multiple instance learning network for whole slide image classification with self-supervised contrastive learning,'' in {\em {IEEE} Conference on Computer Vision and Pattern Recognition, {CVPR} 2021, virtual, June 19-25, 2021}, pp.~14318--14328, Computer Vision Foundation / {IEEE}, 2021.

\bibitem{abmil}
M.~Ilse, J.~M. Tomczak, and M.~Welling, ``Attention-based deep multiple instance learning,'' in {\em Proceedings of the 35th International Conference on Machine Learning, {ICML} 2018, Stockholmsm{\"{a}}ssan, Stockholm, Sweden, July 10-15, 2018} (J.~G. Dy and A.~Krause, eds.), vol.~80 of {\em Proceedings of Machine Learning Research}, pp.~2132--2141, {PMLR}, 2018.

\bibitem{dgmil}
L.~Qu, X.~Luo, S.~Liu, M.~Wang, and Z.~Song, ``{DGMIL:} distribution guided multiple instance learning for whole slide image classification,'' in {\em Medical Image Computing and Computer Assisted Intervention - {MICCAI} 2022 - 25th International Conference, Singapore, September 18-22, 2022, Proceedings, Part {II}} (L.~Wang, Q.~Dou, P.~T. Fletcher, S.~Speidel, and S.~Li, eds.), vol.~13432 of {\em Lecture Notes in Computer Science}, pp.~24--34, Springer, 2022.

\bibitem{yang2021causal}
X.~Yang, H.~Zhang, G.~Qi, and J.~Cai, ``Causal attention for vision-language tasks,'' in {\em Proceedings of the IEEE/CVF Conference on Computer Vision and Pattern Recognition}, pp.~9847--9857, 2021.

\bibitem{li2024causality}
X.~Li, R.~Guo, H.~Zhu, T.~Chen, and X.~Qian, ``A causality-informed graph intervention model for pancreatic cancer early diagnosis,'' {\em IEEE Transactions on Artificial Intelligence}, 2024.

\bibitem{wang2021causal}
T.~Wang, C.~Zhou, Q.~Sun, and H.~Zhang, ``Causal attention for unbiased visual recognition,'' in {\em Proceedings of the IEEE/CVF International Conference on Computer Vision}, pp.~3091--3100, 2021.

\bibitem{ding2022word}
L.~Ding, D.~Yu, J.~Xie, W.~Guo, S.~Hu, M.~Liu, L.~Kong, H.~Dai, Y.~Bao, and B.~Jiang, ``Word embeddings via causal inference: Gender bias reducing and semantic information preservation,'' in {\em Proceedings of the AAAI Conference on Artificial Intelligence}, vol.~36, pp.~11864--11872, 2022.

\bibitem{yao2023interventional}
J.~Yao, X.~Zhu, J.~Jiao, X.~Li, {\em et~al.}, ``Interventional bag multiple instance learning on whole-slide pathological images,'' in {\em Proceedings of the IEEE/CVF Conference on Computer Vision and Pattern Recognition (CVPR)}, pp.~15993--16003, 2023.

\bibitem{chen2025multi}
R.~J. Chen, T.~Ding, M.~Y. Lu, {\em et~al.}, ``Multimodal foundation models for computational pathology,'' {\em Nature Medicine}, vol.~31, no.~2, pp.~243--256, 2025.

\bibitem{zhang2024camil}
Y.~Zhang, H.~Wang, C.~Li, X.~Qian, {\em et~al.}, ``Camil: Context-aware multiple instance learning for whole slide image classification,'' {\em IEEE Transactions on Medical Imaging}, vol.~43, no.~8, pp.~2891--2903, 2024.

\bibitem{seyyed2024demographic}
L.~Seyyed-Kalantari, H.~Zhang, M.~McDermott, I.~Y. Chen, and M.~Ghassemi, ``Demographic-aware medical image analysis: Integrating fairness in deep learning models,'' in {\em Medical Image Computing and Computer Assisted Intervention (MICCAI)}, pp.~412--422, Springer, 2024.

\bibitem{lin2024improving}
M.~Lin, T.~Li, Z.~Sun, G.~Holste, Y.~Ding, F.~Wang, G.~Shih, and Y.~Peng, ``Improving fairness of automated chest radiograph diagnosis by contrastive learning,'' {\em Radiology: Artificial Intelligence}, vol.~6, no.~5, p.~e230342, 2024.

\bibitem{lin2023improving}
M.~Lin, T.~Li, Y.~Yang, G.~Holste, Y.~Ding, S.~H. Van~Tassel, K.~Kovacs, G.~Shih, Z.~Wang, Z.~Lu, {\em et~al.}, ``Improving model fairness in image-based computer-aided diagnosis,'' {\em Nature communications}, vol.~14, no.~1, p.~6261, 2023.

\bibitem{lin2023evaluate}
M.~Lin, Y.~Xiao, B.~Hou, T.~Wanyan, M.~M. Sharma, Z.~Wang, F.~Wang, S.~Van~Tassel, and Y.~Peng, ``Evaluate underdiagnosis and overdiagnosis bias of deep learning model on primary open-angle glaucoma diagnosis in under-served populations,'' {\em AMIA Summits on Translational Science Proceedings}, vol.~2023, p.~370, 2023.

\bibitem{ilse2018attention}
M.~Ilse, J.~M. Tomczak, and M.~Welling, ``Attention-based deep multiple instance learning,'' in {\em International Conference on Machine Learning}, pp.~2127--2136, 2018.

\bibitem{li2021dual}
B.~Li, Y.~Li, and K.~W. Eliceiri, ``Dual-stream multiple instance learning network for whole slide image classification with self-supervised contrastive learning,'' in {\em Proceedings of the IEEE/CVF Conference on Computer Vision and Pattern Recognition}, pp.~14318--14328, 2021.

\bibitem{transmil}
Z.~Shao, H.~Bian, Y.~Chen, Y.~Wang, J.~Zhang, X.~Ji, {\em et~al.}, ``Transmil: Transformer based correlated multiple instance learning for whole slide image classification,'' {\em Advances in Neural Information Processing Systems}, vol.~34, pp.~2136--2147, 2021.

\bibitem{vaswani2017attention}
A.~Vaswani, N.~Shazeer, N.~Parmar, J.~Uszkoreit, L.~Jones, A.~N. Gomez, L.~Kaiser, and I.~Polosukhin, ``Attention is all you need,'' {\em Advances in Neural Information Processing Systems}, vol.~30, 2017.

\bibitem{lu2021data}
M.~Y. Lu, D.~F. Williamson, T.~Y. Chen, R.~J. Chen, M.~Barbieri, and F.~Mahmood, ``Data-efficient and weakly supervised computational pathology on whole-slide images,'' {\em Nature Biomedical Engineering}, vol.~5, no.~6, pp.~555--570, 2021.

\bibitem{pearl2016causal}
J.~Pearl, M.~Glymour, and N.~P. Jewell, {\em Causal Inference in Statistics: A Primer}.
\newblock Wiley, 2016.

\bibitem{pearl2014interpretation}
J.~Pearl, ``Interpretation and identification of causal mediation,'' {\em Psychological Methods}, vol.~19, p.~459, 2014.

\bibitem{stanley2022fairness}
E.~A. Stanley, M.~Wilms, P.~Mouches, and N.~D. Forkert, ``Fairness-related performance and explainability effects in deep learning models for brain image analysis,'' {\em Journal of Medical Imaging}, vol.~9, p.~061102, 2022.

\bibitem{seyyed2021underdiagnosis}
L.~Seyyed-Kalantari, H.~Zhang, M.~B. McDermott, I.~Y. Chen, and M.~Ghassemi, ``Underdiagnosis bias of artificial intelligence algorithms applied to chest radiographs in under-served patient populations,'' {\em Nature Medicine}, vol.~27, pp.~2176--2182, 2021.

\bibitem{beauchamp2003methods}
T.~L. Beauchamp, ``Methods and principles in biomedical ethics,'' {\em Journal of Medical Ethics}, vol.~29, pp.~269--274, 2003.

\bibitem{liu2023translational}
M.~Liu {\em et~al.}, ``A translational perspective towards clinical ai fairness,'' {\em NPJ Digital Medicine}, vol.~6, p.~172, 2023.

\bibitem{srivastava2019mathematical}
M.~Srivastava, H.~Heidari, and A.~Krause, ``Mathematical notions vs. human perception of fairness: a descriptive approach to fairness for machine learning,'' in {\em Proceedings of the 25th ACM SIGKDD International Conference on Knowledge Discovery \& Data Mining}, pp.~2459--2468, ACM, 2019.

\bibitem{jones2010building}
N.~Jones, C.~Bromley, C.~Creegan, R.~Kinsella, F.~Dobbie, and R.~Ormston, ``Building understanding of fairness, equality and good relations,'' Tech. Rep. Research Report 53, Equality and Human Rights Commission, 2015.

\bibitem{amores2013multiple}
J.~Amores, ``Multiple instance classification: Review, taxonomy and comparative study,'' {\em Artificial Intelligence}, vol.~201, pp.~81--105, 2013.

\bibitem{bejnordi2017diagnostic}
B.~E. Bejnordi, M.~Veta, P.~J. Van~Diest, B.~Van~Ginneken, N.~Karssemeijer, G.~Litjens, J.~A. Van Der~Laak, M.~Hermsen, Q.~F. Manson, M.~Balkenhol, {\em et~al.}, ``Diagnostic assessment of deep learning algorithms for detection of lymph node metastases in women with breast cancer,'' {\em Jama}, vol.~318, no.~22, pp.~2199--2210, 2017.

\bibitem{tcga}
L.~A. Cooper, E.~G. Demicco, J.~H. Saltz, R.~T. Powell, A.~Rao, and A.~J. Lazar, ``Pancancer insights from the cancer genome atlas: the pathologist's perspective,'' {\em The Journal of pathology}, vol.~244, no.~5, pp.~512--524, 2018.

\bibitem{kang2023benchmarking}
M.~Kang, H.~Song, S.~Park, D.~Yoo, and S.~Pereira, ``Benchmarking self-supervised learning on diverse pathology datasets,'' in {\em Proceedings of the IEEE/CVF Conference on Computer Vision and Pattern Recognition}, pp.~3344--3354, 2023.

\bibitem{marugoto}
K.~Lab, ``marugoto: Machine learning for medical images,'' 2024.

\bibitem{loshchilov2016sgdr}
I.~Loshchilov and F.~Hutter, ``Sgdr: Stochastic gradient descent with warm restarts,'' in {\em Proceedings of the 5th International Conference on Learning Representations (ICLR)}, 2017.

\bibitem{dosovitskiy2021image}
A.~Dosovitskiy, L.~Beyer, A.~Kolesnikov, D.~Weissenborn, X.~Zhai, T.~Unterthiner, M.~Dehghani, M.~Minderer, G.~Heigold, S.~Gelly, J.~Uszkoreit, and N.~Houlsby, ``An image is worth 16x16 words: Transformers for image recognition at scale,'' in {\em International Conference on Learning Representations (ICLR)}, 2021.

\bibitem{wang2022transformer}
X.~Wang, S.~Yang, J.~Zhang, M.~Wang, J.~Zhang, W.~Yang, J.~Huang, and X.~Han, ``Transformer-based unsupervised contrastive learning for histopathological image classification,'' {\em Medical Image Analysis}, vol.~81, p.~102559, 2022.

\bibitem{chen2024uni}
R.~J. Chen, T.~Ding, M.~Y. Lu, D.~F. Williamson, G.~Jaume, A.~H. Song, B.~Chen, A.~Zhang, D.~Shao, M.~Shaban, {\em et~al.}, ``Towards a general-purpose foundation model for computational pathology,'' in {\em Nature Medicine}, vol.~30, pp.~850--862, Nature Publishing Group, 2024.

\bibitem{oquab2023dinov2}
M.~Oquab, T.~Darcet, T.~Moutakanni, H.~Vo, M.~Szafraniec, V.~Khalidov, P.~Fernandez, D.~Haziza, F.~Massa, A.~El-Nouby, M.~Assran, N.~Ballas, W.~Galuba, R.~Howes, P.-Y. Huang, S.-W. Li, I.~Misra, M.~Rabbat, V.~Sharma, G.~Synnaeve, H.~Xu, H.~Jegou, J.~Mairal, P.~Labatut, A.~Joulin, and P.~Bojanowski, ``Dinov2: Learning robust visual features without supervision,'' in {\em Proceedings of the IEEE/CVF International Conference on Computer Vision (ICCV)}, pp.~3038--3048, 2023.

\bibitem{DTFD-MIL}
H.~Zhang, Y.~Meng, Y.~Zhao, Y.~Qiao, X.~Yang, S.~Coupland, and Y.~Zheng, ``Dtfd-mil: Double-tier feature distillation multiple instance learning for histopathology whole slide image classification,'' in {\em Proceedings of the IEEE/CVF Conference on Computer Vision and Pattern Recognition}, pp.~18802--18812, 2022.

\bibitem{royston2013external}
P.~Royston and D.~G. Altman, ``External validation of a cox prognostic model: Principles and methods,'' {\em BMC Medical Research Methodology}, vol.~13, no.~1, pp.~1--15, 2013.

\bibitem{travis2015lung}
W.~D. Travis, E.~Brambilla, A.~G. Nicholson, Y.~Yatabe, J.~H. Austin, M.~B. Beasley, L.~R. Chirieac, S.~Dacic, E.~Duhig, D.~B. Flieder, K.~Geisinger, F.~R. Hirsch, Y.~Ishikawa, K.~M. Kerr, M.~Noguchi, G.~Pelosi, C.~A. Powell, M.~S. Tsao, I.~Wistuba, W.~Panel, {\em et~al.}, ``The 2015 world health organization classification of lung tumors: Impact of genetic, clinical and radiologic advances since the 2004 classification,'' {\em Journal of Thoracic Oncology}, vol.~10, no.~9, pp.~1243--1260, 2015.

\end{thebibliography}
\end{document}